\documentclass[final,5p,times,twocolumn]{elsarticle}

\usepackage{lineno,hyperref}
\modulolinenumbers[5]

\journal{ISPRS Journal of Photogrammetry and Remote Sensing}

\usepackage{tikz}
\usepackage{xcolor}
\usepackage{color}

\usepackage[binary-units = true]{siunitx}
\DeclareSIUnit\px{px}
\usepackage{subcaption}
\usepackage{amsfonts} 
\usepackage{pgf}
\usepackage[utf8]{inputenc}\DeclareUnicodeCharacter{2212}{-}
\usepackage{booktabs}

\begin{document}

\begin{frontmatter}

\title{Artificial and beneficial -- Exploiting artificial images for aerial vehicle detection}

\author[1]{Immanuel Weber\corref{cor1}}
\ead{immanuel.weber@hs-koblenz.de}
\author[1]{Jens Bongartz}
\ead{bongartz@hs-koblenz.de}
\author[2]{Ribana Roscher}
\ead{ribana.roscher@uni-bonn.de}

\cortext[cor1]{Corresponding author}
\address[1]{Application Center for Machine Learning and Sensor Technology, University of Applied Sciences Koblenz, 53424 Remagen,
Germany}
\address[2]{Institute of Geodesy and Geoinformation, University of Bonn, 53115 Bonn, 
Germany}

\begin{abstract}
Object detection in aerial images is an important task in environmental, economic, and infrastructure-related tasks.
One of the most prominent applications is the detection of vehicles, for which deep learning approaches are increasingly used.
A major challenge in such approaches is the limited amount of data that arises, for example, when more specialized and rarer vehicles such as agricultural machinery or construction vehicles are to be detected.
This lack of data contrasts with the enormous data hunger of deep learning methods in general and object recognition in particular.
In this article, we address this issue in the context of the detection of road vehicles in aerial images.
To overcome the lack of annotated data, we propose a generative approach that generates top-down images by overlaying artificial vehicles created from 2D CAD drawings on artificial or real backgrounds.
Our experiments with a modified RetinaNet object detection network show that adding these images to small real-world datasets significantly improves detection performance. 
In cases of very limited or even no real-world images, we observe an improvement in average precision of up to 0.70 points.
We address the remaining performance gap to real-world datasets by analyzing the effect of the image composition of background and objects and give insights into the importance of background.
\end{abstract}


\end{frontmatter}


\section{Introduction}
\label{sec:introduction}

Object detection in aerial images plays a significant role in remote sensing applications like traffic monitoring, infrastructure surveillance, and environmental protection~\cite{heipke2020deep,kaack2019truck,ma2019deep,cazzato2020survey,torney2019comparison}.
Modern object detectors, based on deep neural networks, like the families of methods evolving around Faster R-CNN~\cite{ren2015faster}, RetinaNet~\cite{lin2017focal} or YOLO~\cite{redmon2015you}, with more recent variants like EfficentDet~\cite{tan2020efficientdet} and YOLOv4~\cite{bochkovskiy2020yolov4}, have proven their capabilities in general benchmarks containing earth-bound imagery \cite{lin2017focal,lin2014microsoft}, and also in specific aerial image benchmarks \cite{sun2018salience,xia2018dota,van2019satellite}.
However, like most machine learning and especially deep learning methods, they generally require vast amounts of data to learn from ~\cite{sun2017revisiting}.
Also, object detection is a supervised task that requires annotated data, which typically involves human labor.
The lack of annotated data is often present and a problem for many real-world applications, and therefore a potential inhibitor for wider acceptance of deep learning solutions~\cite{ma2019deep}.

For tasks like vehicle detection in aerial images, the challenge is particularly apparent if certain types of vehicles are to be identified.
Focusing on road vehicles and considering the distribution of specific vehicle types in the real world, we see that cars and trucks occur most often. 
In contrast, more specialized vehicles like agricultural machinery, construction vehicles, or municipal vehicles are scarce.
Even without a finer class subdivision than these rough categories, it is clear that an unbalanced long-tail distribution is present.
Besides the scarcity of specific classes, we are also challenged by a large intra-class variety due to different vehicle colors, shapes, and varying scenery resulting from the different landscape, background structures, weather, and various lighting conditions.
As a result, collecting data covering the whole intra-class variety is associated with intensive acquisition and annotation work. 
To overcome this challenge, much work is put into topics targeting a lack of data from different perspectives \cite{takahashi2019data,sun2016deep,xie2020self,peng2015learning,koga2020method,zheng2019using}.
A particularly promising approach is the generation of annotated data, which enables a flexible creation of data samples for classes underrepresented or missing.

In this paper, we detect vehicles in aerial images, where a generative approach is utilized to create artificial images in order to overcome the challenge of limited annotated data.
Our main objective is to provide a better understanding of the effect of a lack of data and to what extent artificial data can compensate this effect.
For this purpose, we utilize and modify the ISPRS Potsdam dataset \cite{rottensteiner2013} that contains a large number of objects and images.
We focus on only one class, \texttt{cars}, that allows us to simulate a lack of data and determine the difference between various training dataset sizes.
In this way, we mimic rare classes such as agricultural machinery or construction vehicles but do not have to constrain our analyses due to lack of data. 
We propose an image generator that uses simplified 2D computer-aided-design (CAD) drawings to create images that boost the model's training.
In order to approach the phenomenon of rare objects even in the actually common vehicle class, we use here only a few 2D CAD drawings. 
The latter also takes into account the possible limited accessibility or diversity of these drawings.
We will show that simple 2D CAD drawings are effective when used alone or in combination with small numbers of real images to train an accurate object detector.  
Moreover, we will demonstrate that the background has a significant impact on the achievable performance, and we investigate further into the interplay of objects and background in object detection.

Our contributions are as follows:
\begin{itemize}
    \item We introduce an image generator for top-down images which uses simplified 2D CAD drawings to create artificial images suitable for training an object detector. The approach can be extended to an arbitrary amount of classes and is especially promising for rare classes. 
    \item We provide an analysis about the correlation of dataset size and object detection performance, and the impact of adding artificial images.
    \item We provide insights into the importance of image composition, especially regarding the background, and point to promising direction to further improve the detection performance. 
\end{itemize}

The paper is structured as follows:
We first give an overview of relevant work and relate our approach to them in Section \ref{sec:related_work}.
We continue by describing our object detector and proposed image generator in Section \ref{sec:framework}.
In Section \ref{sec:data} we introduce the real-world dataset which is used for our experiments. 
We evaluate our artificial imagery's effect on our task and analyze it in detail in Section \ref{sec:experiments}.
Section \ref{sec:conclusion_future} closes this work with our final conclusions and directions for future work.

\section{Related work}
\label{sec:related_work}

As indicated initially, the lack of data is a core problem of many deep learning applications.
Since data shortage has different aspects, various approaches have been proposed to overcome this problem.
In the following, we group relevant works and discuss their intended use.
Moreover, we position our work and relate it to the discussed approaches.

\paragraph{Image augmentation}
Simple methods like flipping, cropping, and color changes of existing images, documented in works like AlexNet~\cite{krizhevsky2017imagenet} and ResNet~\cite{he2016deep} are commonly used nowadays and are required to let models generalize well even on datasets of the size of ImageNet \cite{deng2009imagenet}.
The random expansion and cropping augmentation used with the Single Shot MultiBox Detector \cite{liu2016ssd}, a predecessor of the RetinaNet, was one of the keys to its success.
Several more complex augmentation methods \cite{takahashi2019data,devries2017improved,yun2019cutmix,zhang2017mixup}, of whom especially mixup~\cite{zhang2019bag} and Random Image Cropping and Patching (RICAP, \cite{takahashi2019data}) are suitable for object detection, highlighted that well-designed augmentations have the potential to improve performance significantly, particularly in the long run. 

\paragraph{Transfer-learning} 
An approach to shorten training times is transfer-learning, which can help jump-start model training using decent pre-trained models that have already captured key vision features.
As a result, trained models generalize better also to smaller datasets~\cite{sharif2014cnn,girshick2013rich}.
However, Shen et al.~\cite{shen2019object} show that this is not always better than training from scratch if one can forgo shorter training times.
It remains an open research question to what extent this holds for small datasets.

\paragraph{Domain adaption}
This group of methods tries to transfer data and knowledge from one domain to another, which is useful if high quality, annotated datasets exist in a domain other than the target domain.
Methods like CORAL \cite{sun2016deep}, which have been successfully applied to aerial object detection~\cite{koga2020method}, or the generative adversarial network-based CycleGAN~\cite{zhu2017unpaired}, work in a semi-supervised manner.
These methods have in common that by transferring the existing images from one to another domain, the annotations can be reused, and hence no new annotation work is required.

\paragraph{Semi-supervised learning}
Most tasks, like image classification, object detection, or semantic segmentation, are supervised tasks.
In contrast to that, semi-supervised learning methods also utilize information from large unlabeled datasets, which has proven successful, especially for small labeled datasets from the same domain.
Methods like MixMatch, FixMatch~\cite{berthelot2019mixmatch,sohn2020fixmatch}, STAC~\cite{sohn2020simple}, SimCLR~\cite{chen2020simple} or Noisy Student~\cite{xie2020self} use labeled data for model training and let them make predictions on unlabeled ones.
Based on different approaches, they then use these predictions to improve the underlying models. 

\paragraph{Artificial image generation}
All of the approaches above have in common that they require existing annotated data.
This is different for artificial image generation methods that can create new images to extend existing datasets or build new datasets from scratch.
They allow the automation of the generation of images and annotations and, therefore, be easier and cheaper than real data acquisition.
Classic computer graphic approaches, like 3D rendering engines \cite{peng2015learning,stark2010back,movshovitz2016useful,tremblay2018training}, but also video games \cite{richter2016playing}, can create photo-realistic images that are useful for a variety of domains including aerial images \cite{shermeyer2020rareplanes}.
Whereas the image generation process can be automated, the underlying artificial scenes' design requires 3D models, textures, and potentially human labor. 
However, it also gives complete control over the content of the images.

\paragraph{Generative adversarial networks}
Generative adversarial networks (GANs, \cite{goodfellow2014generative}) are less a separate group but can be used in most of the presented approaches.
They have proven to be powerful \cite{zheng2019using,shrivastava2017learning,park2019semantic}, however, 
as they are a learning-based approach, they also require relevant data in a sufficient amount and are bound to the distribution of that data.

\paragraph{Further approaches}
Further methods exist that address other aspects that are helpful for data shortage.
Examples are image harmonization using deep image priors~\cite{ulyanov2018deep,bhattad2020cut} or few-shot learning methods~\cite{xian2018zero}, that try to reduce the amount of data needed for training to a minimum.
However, new data, outside of existing distributions, can only be created with the rendering methods described before or other methods that can introduce out-of-distribution content.

\paragraph{Relation to our approach}
None of these approaches will probably solve the issue of lacking data alone, but a combination can help achieve its intended success.
For our situation, in which many rare classes may exist, and no access to large amounts of data is available, we consider the generation of artificial images as the most promising approach that, however, can still be combined with other methods.
Once a suitable number of images with objects from the required distribution is available, GAN-based image refinement~\cite{shrivastava2017learning} and image augmentation~\cite{zheng2019using}, image harmonization~\cite{bhattad2020cut} and domain adaption \cite{koga2020method} can come into play to be applied on top of these artificial images to bridge gaps between these and real images.

\section{Framework}
\label{sec:framework}
In the following, we will introduce our framework, which consists of two main components: an object detector and an image generator.
Most object detection networks, such as the RetinaNet that we have chosen, are in principle suitable for our task.
However, we modified the network to improve its performance for our application, which is described in section \ref{sec_sub:framework-detector}.
In section \ref{sec_sub:framework-art}, we present the image generation process and the underlying data.

\subsection{Object detector}
\label{sec_sub:framework-detector}

In our framework, we aim for an object detector designed to perform well on aerial images.
Although many object detectors specialized for aerial images have been proposed in the past (see, e.g., \cite{cazzato2020survey}, \cite{ophoff2020vehicle}), we observe that standard object detectors with small modifications are suitable for our data.
We decide to use the RetinaNet~\cite{lin2017focal} for this work and modify it accordingly.
Unlike networks of the Faster R-CNN family~\cite{ren2015faster}, RetinaNet is a single-step process, i.e., it does not require an intermediate step like region proposal networks and therefore can be trained end-to-end.
It does not use direct regression to locate objects but instead uses support points distributed throughout the image, called anchors, to regress the position and size of detections relative to these anchors.
The vanilla RetinaNet is designed for common datasets like MS COCO~\cite{lin2014microsoft}, which typically contain earthbound imagery.
In contrast, our aerial images, for example, have a different perspective and scale \cite{cazzato2020survey}, which we take into account with the following modifications.

\begin{figure}[!ht]
    \centering
    \includegraphics[width=\linewidth]{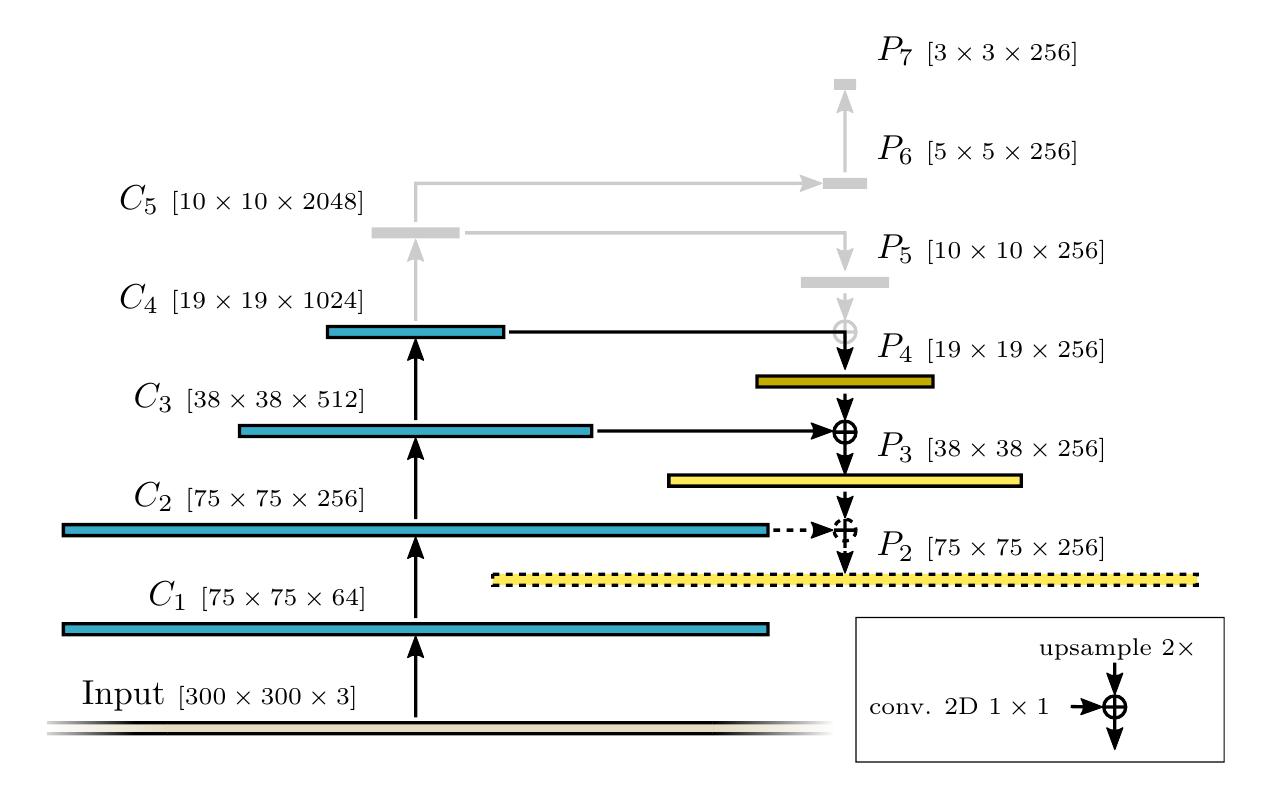}
    \caption{Modified feature pyramid -- The left, blue feature maps, $C_{1}$ to $C_{5}$, are computed with the ResNet feature extractor's convolutional blocks.
    The right, yellow stack is the actual feature pyramid with feature maps $P_{2}$ to $P_{7}$.
    $P_{5}$ to $P_{6}$ are computed using standard 2D convolutions that reduce spatial or depth extent.
    The bottom right box shows the computation of the $P_{i}$.
    The gray-colored feature maps are the ones we removed.
    The dashed lines mark our extensions, and the darker yellow feature map is used for feature map computation but not for the subsequent prediction.
    The sizes of the feature maps, also provided in the braces, match proportionally, but the input is only partially visible.
    } 
    \label{fig:framework-fpn}
\end{figure}

The detector's core is a feature pyramid network~\cite{lin2017feature} that is built on a feature extraction network, like a ResNet-50~\cite{he2016deep}.
Such a network usually consists of several convolutional blocks, which in sequential application reduce the spatial resolution of an input and increase the receptive field accordingly.
The basic idea of feature pyramid networks is to split these feature extraction networks and collect the feature maps computed by each convolutional block.
This gives us access to a set of feature maps with different receptive fields that can be used to detect objects of different sizes.
As these feature maps become spatially progressively smaller, they form the pyramid with respective denotation.
This pyramid for a ResNet with its feature maps $C_{1}$ to $C_{5}$ is shown in Figure \ref{fig:framework-fpn} on the left.
By applying convolution and up-sampling steps, as shown in the figure on the right, these feature maps, $P_{2}$ to $P_{7}$, are enriched to transfer information from higher to lower pyramid feature maps.
The vanilla RetinaNet has a feature pyramid with the five feature maps $P_{3}$ to $P_{7}$.
Due to the relatively small size of the objects in our data, most of the higher feature maps do not contribute to the detection task.
This is expected, as the first convolutional block of the feature extractor network already has a feature map of only a fourth of the input size. Each of the following blocks reduces it further by a factor of two.
Therefore, these feature maps have an increasingly larger receptive field and activate on larger objects and have coarser localization abilities.
As shown in Figure \ref{fig:framework-fpn}, we remove the top three feature maps $P_{5}$ to $P_{7}$ but add a new feature map $P_{2}$ to the bottom, which works on the largest available feature map size and thus, is having a smaller receptive field but is also able to finer localize small objects.
Also, we do not use $P_{4}$ for predictions and only use it for the top-down enrichment of $P_{3}$ and $P_{2}$ downstream.
The final feature maps used for prediction, $P_{2}$ and $P_{3}$, much more resemble a very shallow frustum than a pyramid but are more efficient and improve performance slightly.

We further add two modifications that were introduced in the works of Fully Convolutional One-Stage Object Detection (FCOS)~\cite{tian2019fcos} and Adaptive Training Sample Selection (ATSS)~\cite{zhang2020bridging}.
The first is a learnable scalar for each pyramid feature map used to scale the bounding box regression branch's logits to better account for different field-of-views per pyramid level.
The second is a new task branch for the prediction of the centerness, which is a measure that describes how well an anchor is centered in its predicted bounding box.
In the inference phase, this measure is used as a weight for the anchors' confidences and results in lower confidences for predicted boxes, which are far away from their anchor's center.
Since the predictions are ordered by their confidence, as a result, these boxes are less likely to persist after the non-maximum-suppression, that is used to remove multiple overlapping predictions.
As the centerness is only defined inside a bounding box, the matching mechanism between ground truth bounding boxes and anchors is slightly modified.
Anchors, which are regarded as positive candidates by the RetinaNet mechanism but whose centers are outside of the ground truth bounding box, are ignored.

\begin{figure}[!ht]
    \centering
    \includegraphics[width=\linewidth]{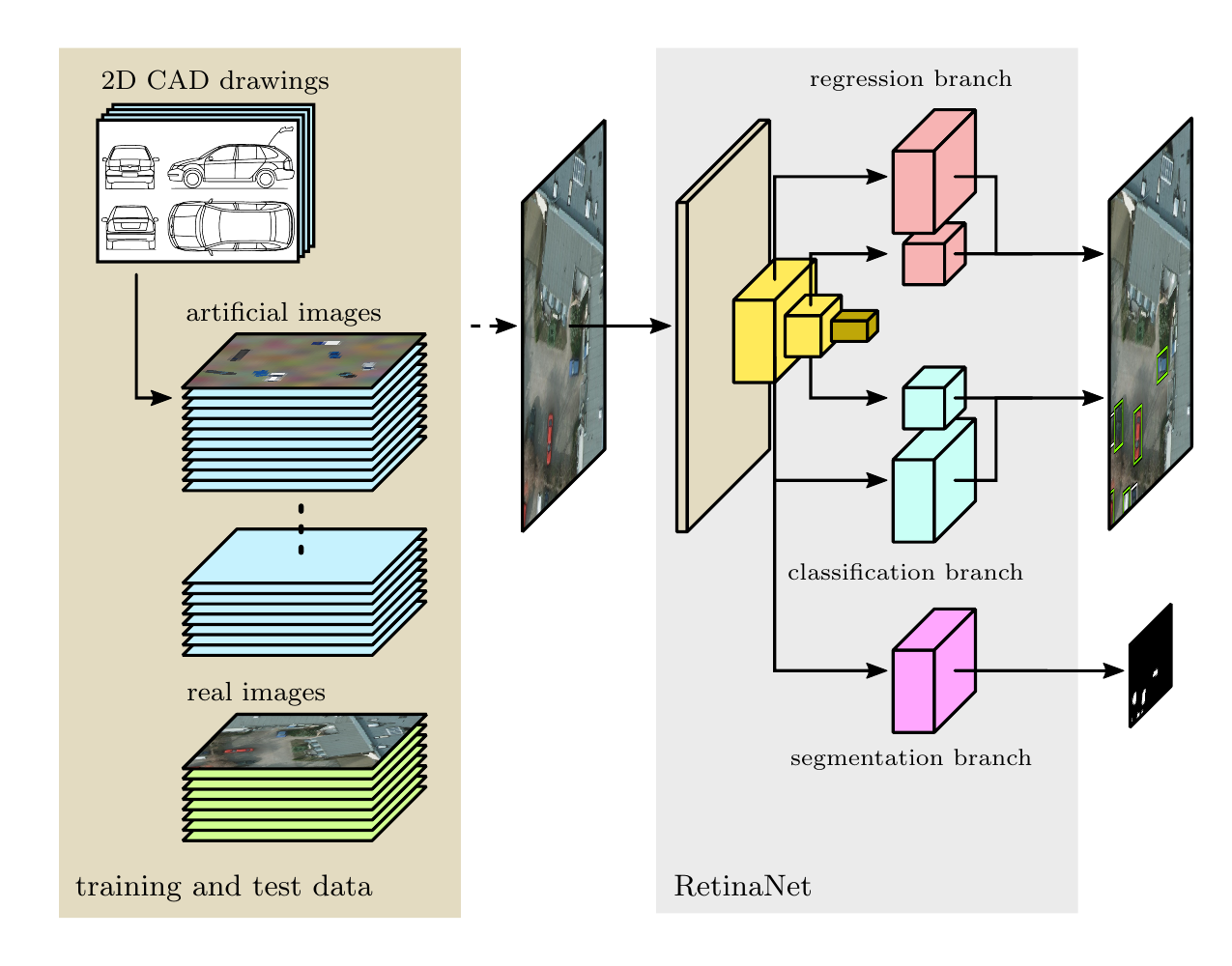}
    \caption{Dataset and object detector structure -- On the left side, the two different data sources, real and artificial images (see Section \ref{sec_sub:framework-art}), are shown. 
    The modified RetinaNet with the shallower three-level feature pyramid network (yellow) and the additional segmentation branch (magenta) are shown on the right side. 
    The outputs for a single sample are shown, and the spatial extents of the feature maps are proportional to their real size and show the spatial reduction in the pyramid. 
    We have omitted the centerness branch and connection details of the feature pyramid for simplicity.} 
    \label{fig:framework-detektor}
\end{figure}

We add a fourth task branch for semantic segmentation, that we will use in Section \ref{sec_sub:art-composition} to gain more insights about the learned representation of our models and which is loosely inspired by J\"ager et al.~\cite{jaeger2020retina}.
It consists of a single convolutional layer applied only to the feature pyramid level $P_{2}$. 
The resulting activation maps have the size 75 $\times$ 75, a channel for each class, and are being sigmoid activated.
We derive the corresponding annotations by rasterizing the object-aligned bounding boxes available in our annotation data. 

The RetinaNet's structure is fully convolutional, which allows it to process input images of any size. 
The vanilla design is laid out for image sizes within 800 \si{\px} to 1333 \si{\px}.
Varying the input size, however, requires to create a new matching anchor grid.
We fix it to 300 $\times$ 300 \si{\px}, which allows us to use a fixed anchor grid and larger batch sizes due to the smaller image size.
Our only modification to the anchors is that we use four times smaller anchor sizes to account for smaller images.
We use the default RetinaNet sub-losses of smooth L$_1$ loss for bounding box regression and focal loss for classification.
The two new task branches' losses are binary cross-entropy for centerness and the averaged dice and binary cross-entropy loss for semantic segmentation.
We weigh all four losses equally in the final loss summation.
Besides, FCOS and ATSS also propose to use a general-intersection-over-union (GIoU) based loss \cite{rezatofighi2019generalized} as a replacement for the smooth L$_1$ loss.
In \cite{zheng2020distance}, a complete-intersection-over-union (CIoU) loss is introduced, which is said to be superior to the GIoU loss.
We evaluated the standard smooth L$_1$, GIoU, and CIoU loss and, on an anecdotal level, report that they show comparable performance, whereas the latter are decreasing much quicker while training.
However, GIoU and CIoU loss did not converge significantly faster and could not beat the smooth L$_1$ loss in the long run.
Finally, the reference implementation of ATSS~\cite{zhang2020bridging} uses the centerness value as a weight for the bounding box loss, which also did not improve the performance in our case.
An overview of the detector's structure is given in Figure \ref{fig:framework-detektor}.

\subsection{Artificial Image Generator}
\label{sec_sub:framework-art}
\begin{figure}
    \centering
    
    \begin{subfigure}[t]{0.32\linewidth}
    \includegraphics[width=\linewidth]{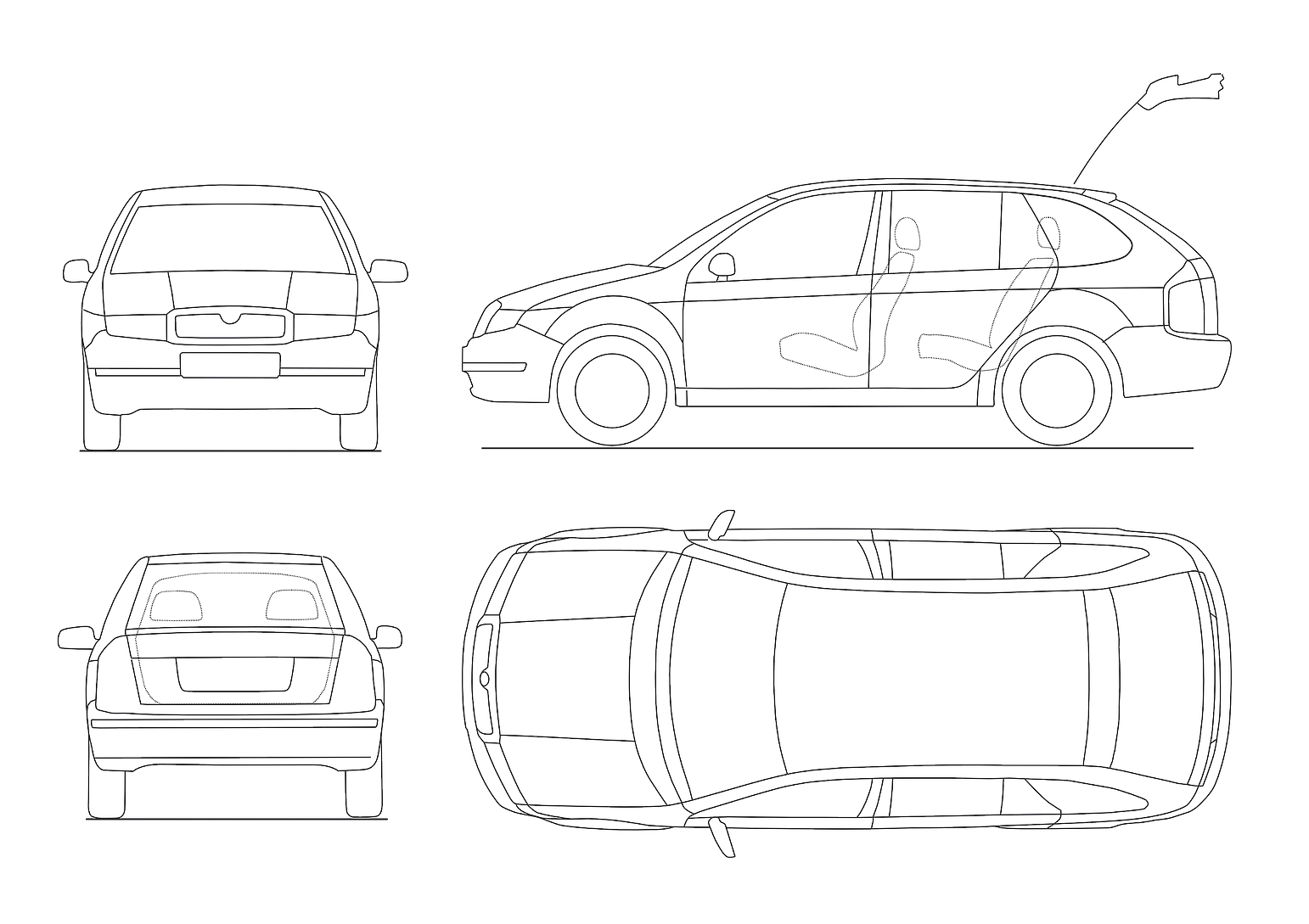}
    \caption{raw simplified drawing} 
    \label{fig:art-gen-blueprints_raw}
    \end{subfigure}
    \hfill
    \begin{subfigure}[t]{0.32\linewidth}
    \includegraphics[width=\linewidth]{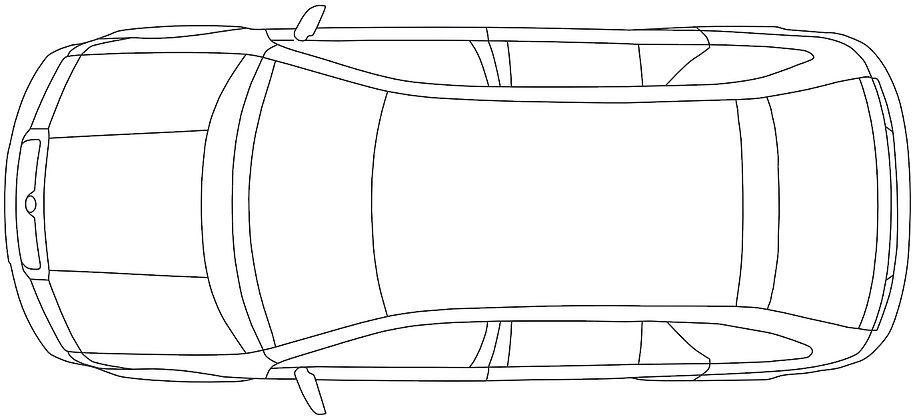}
    \caption{cleaned top-down drawing} 
    \label{fig:art-gen-blueprints_cleaned}
    \end{subfigure}
    \hfill
    \begin{subfigure}[t]{0.32\linewidth}
    \includegraphics[width=\linewidth]{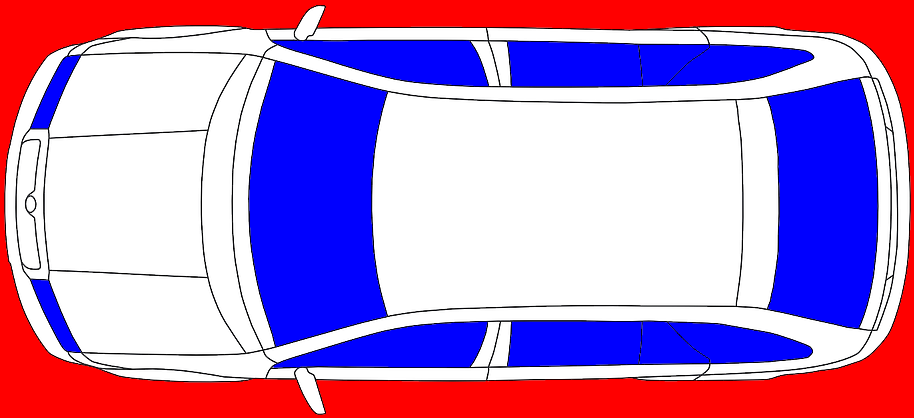}
    \caption{masked drawing} 
    \label{fig:art-gen-blueprints_masked}
    \end{subfigure}
    
    \caption{Preparation of a simplified 2D CAD drawing -- The raw drawing (left) is freed from size markings and irrelevant structures and reduced to the top view (center) and then masked using distinct colors for the surface types (right).}
    \label{fig:art-gen-blueprints}
\end{figure}

In our framework, we design an artificial image generator that is designed under the premise of creating images that support the training and, at the same time, be as simple as possible.
We intend to train models that generalize to details and variations, especially inside a class, so we do not aim to create images of high resolution or exact replicas of relevant objects.
Unlike earthbound images, aerial images show scenes from a bird's eye view so that the majority of the information can be found in two dimensions. 
We take advantage of this and build a simple image generator without 3D models and rendering.
Instead, we use simplified 2D CAD drawings, which can be found, for example, in advertising brochures or maintenance materials of vehicle manufacturers, of which some are available in online databases\footnote{see for example https://drawingdatabase.com/}.
These drawings or blueprints show the vehicle's shape and essential structure, but they are not directly suitable for digital design or rendering due to their actual intention.
We remove details and elements that are not part of the vehicle, such as size markings, and manually create masks for background, outline, body, lights, and windows by coloring them with different colors.
An example of this can be seen in Figure \ref{fig:art-gen-blueprints}.
Our generator reads the processed images, slightly simplifies them to remove further details, and re-scales them to the intended ground sampling distance.

The first step of the actual image generation process is the background canvas creation.
We create a plain background of size 600 $\times$ 600 \si{\px} and color it in the average color of the Imagenet dataset.
Since this color is also used to standard-normalize the images during training, it can be considered a neutral color.
We add a rough and fine noise pattern, which both are generated from a normal distribution.
The rough noise is created on a smaller grid of size 10 $\times$ 10 with $\sigma^2 = 10$, is up-sampled to the background-size using bicubic interpolation, and added to the background canvas, which results in a smooth large scale undulation.
The fine noise is applied per pixel with a $\sigma^2 = 5$.

In order to add objects to the background images, several blueprints are selected randomly, and each of them is colored with random colors considering the surface type, like standard car colors for the body and blueish colors for the windows.
The generator draws the vehicle outlines by default black but can optionally draw it in different colors, e.g., body color.
Some vehicles are randomly cut to 50~\% to 70~\% of the width or height of the instance, as in reality, they may only be partially visible.

\begin{figure}
    \centering
    
    \begin{subfigure}[t]{0.32\linewidth}
    \includegraphics[width=\linewidth]{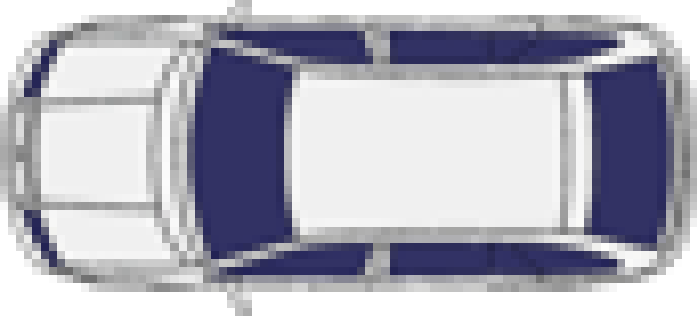}
    \caption{Colored instance.} 
    \end{subfigure}
    \hfill
    \begin{subfigure}[t]{0.32\linewidth}
    \includegraphics[width=\linewidth]{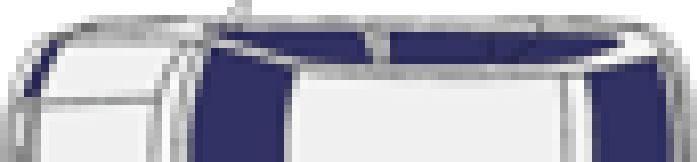}
    \caption{Cut instance.} 
    \label{fig:art-gen-instancing_cutted}
    \end{subfigure}
    \hfill
    \begin{subfigure}[t]{0.32\linewidth}
    \includegraphics[width=\linewidth]{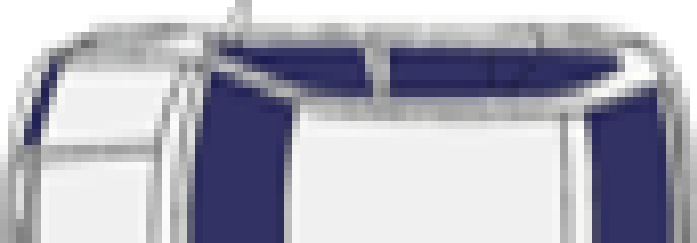}
    \caption{Deformed instance.} 
    \label{fig:art-gen-instancing_deformed}
    \end{subfigure}
    
    \caption{Creation of a vehicle instance -- The blueprint is first painted using suitable colors (left), then optionally cut (center) and deformed (right). For visualization, the object is deformed by 50~\%.}
    \label{fig:art-gen-instancing}
\end{figure}

\begin{figure}
    \centering
    
    \begin{subfigure}[t]{0.32\linewidth}
    \includegraphics[width=\linewidth]{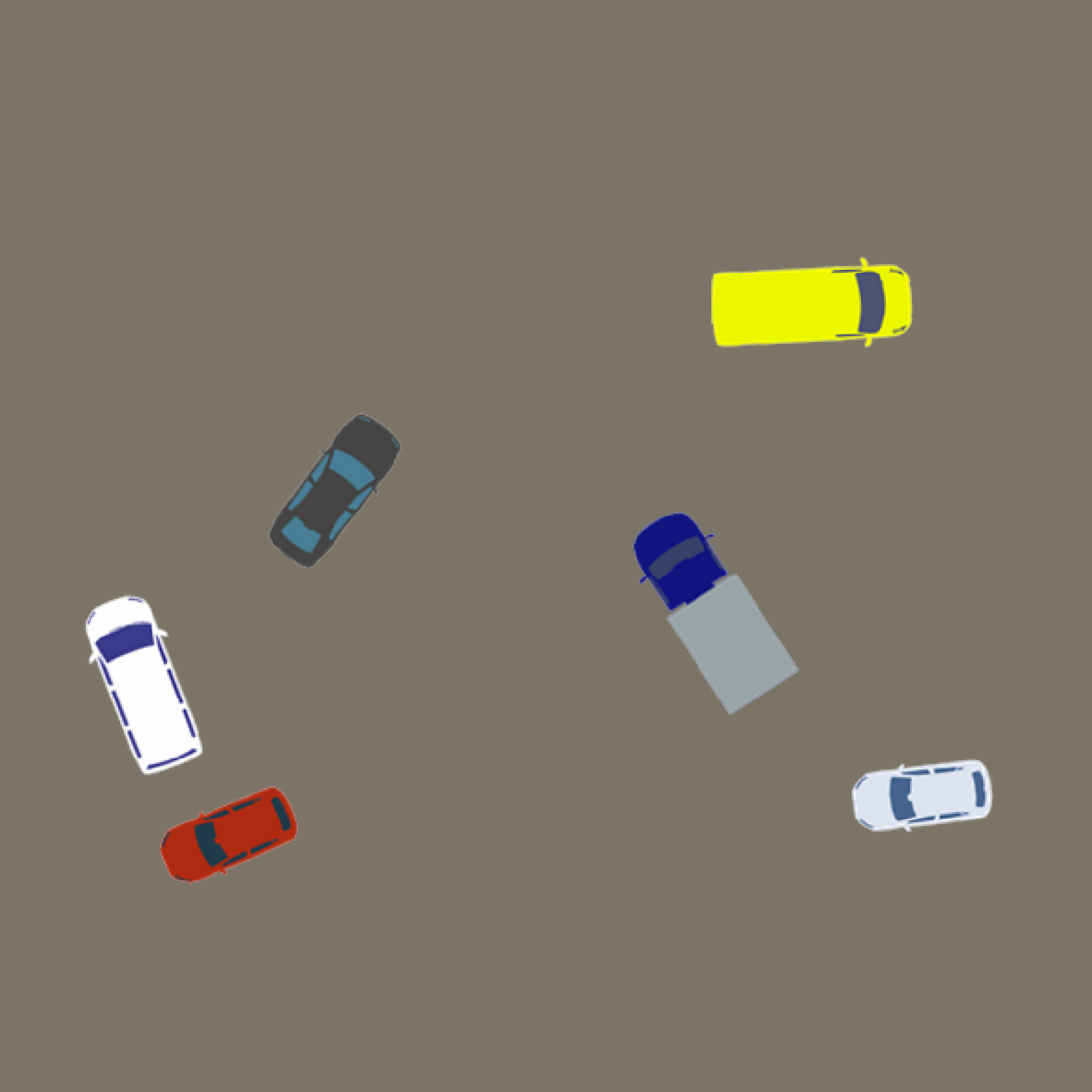}
    \caption{\raggedright plain colored vehicles} 
    \label{fig:art-gen-pipeline_plain}
    \end{subfigure}
    \hfill
    \begin{subfigure}[t]{0.32\linewidth}
    \includegraphics[width=\linewidth]{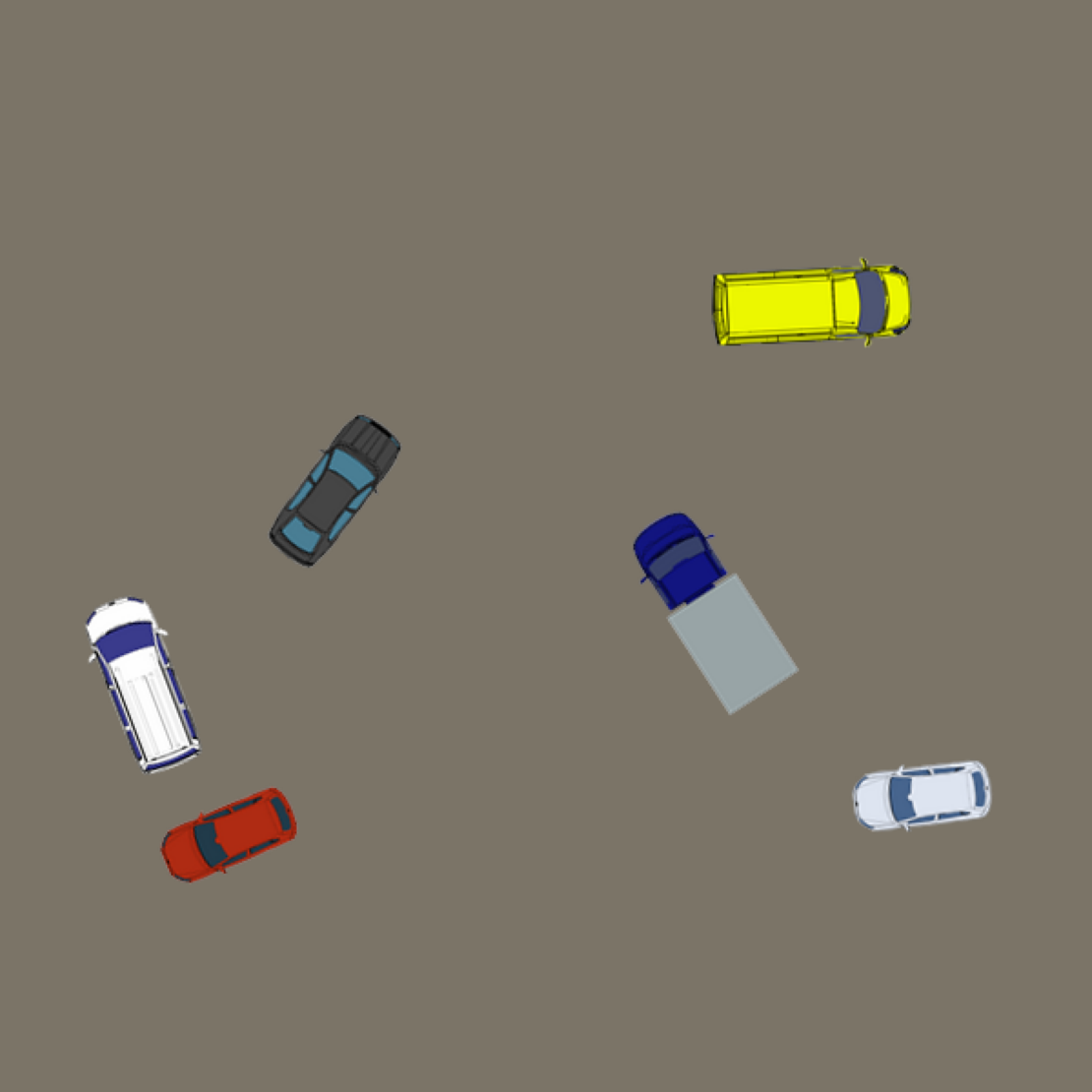}
    \caption{\raggedright + black vehicle outlines} 
    \label{fig:art-gen-pipeline_black_border}
    \end{subfigure}
    \hfill
    \begin{subfigure}[t]{0.32\linewidth}
    \includegraphics[width=\linewidth]{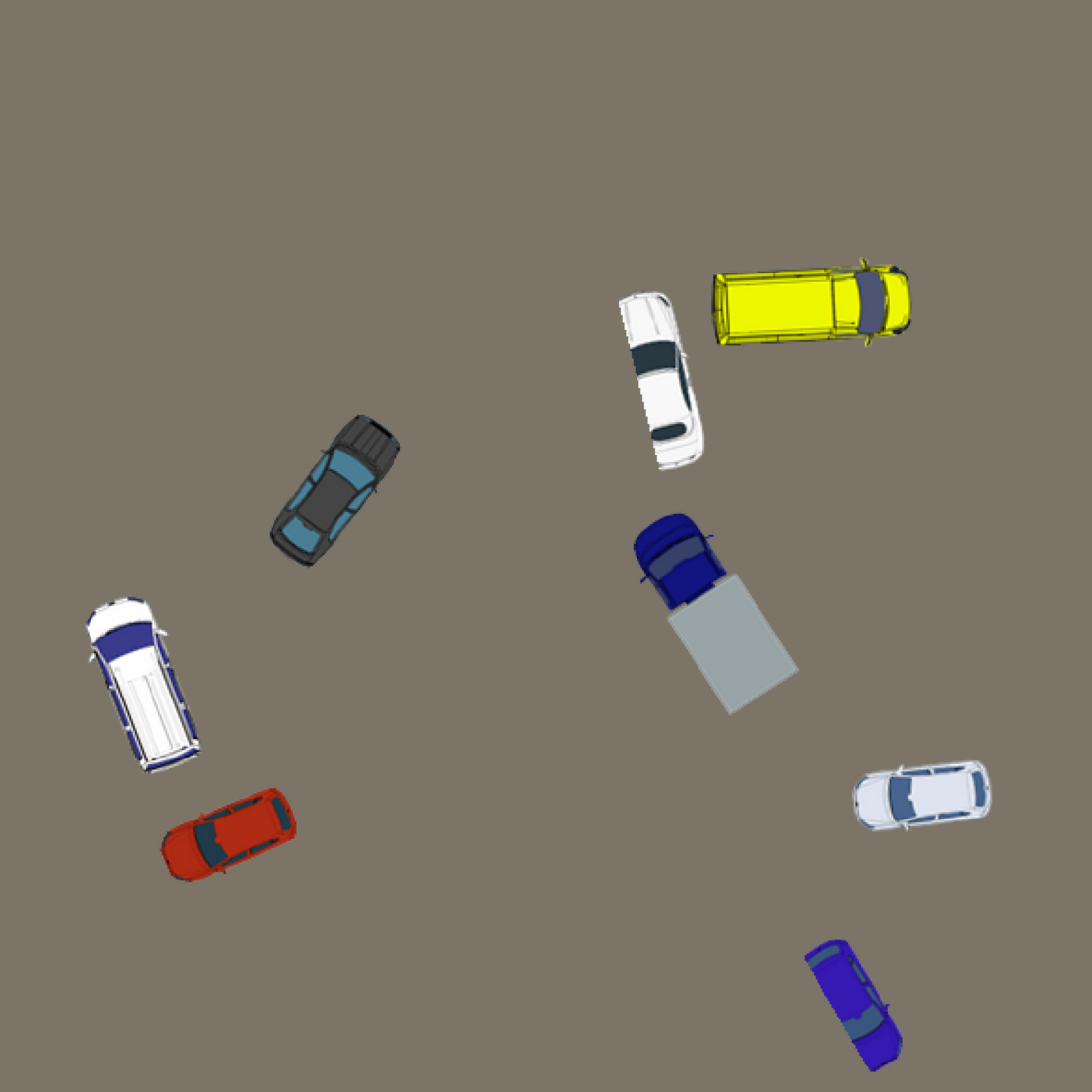}
    \caption{\raggedright + partially depicted vehicles} 
    \label{fig:art-gen-pipeline_parts}
    \end{subfigure}
    
    \medskip
    
    \begin{subfigure}[t]{0.32\linewidth}
    \includegraphics[width=\linewidth]{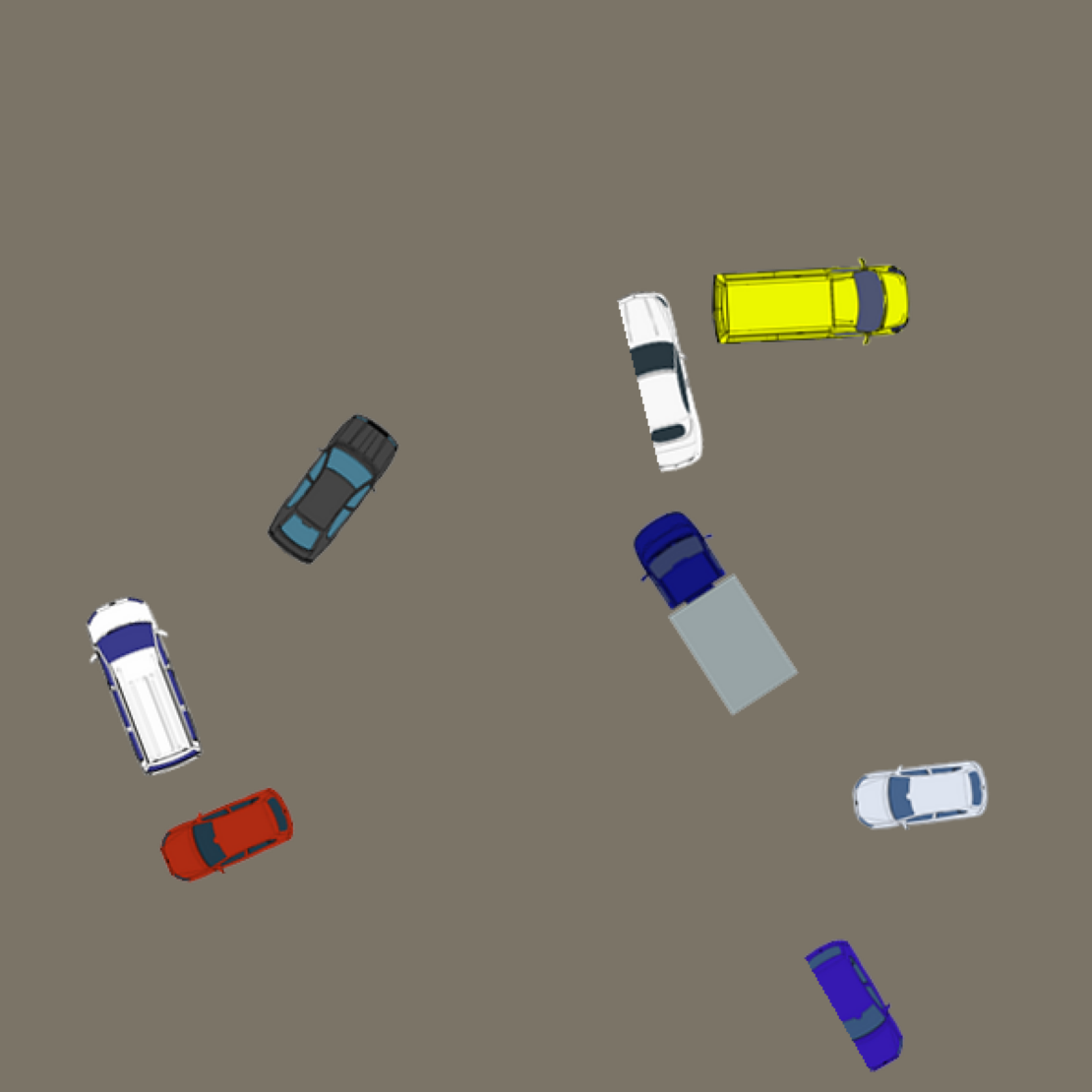}
    \caption{\raggedright + deformations} 
    \label{fig:art-gen-pipeline_deform}
    \end{subfigure}
    \hfill
    \begin{subfigure}[t]{0.32\linewidth}
    \includegraphics[width=\linewidth]{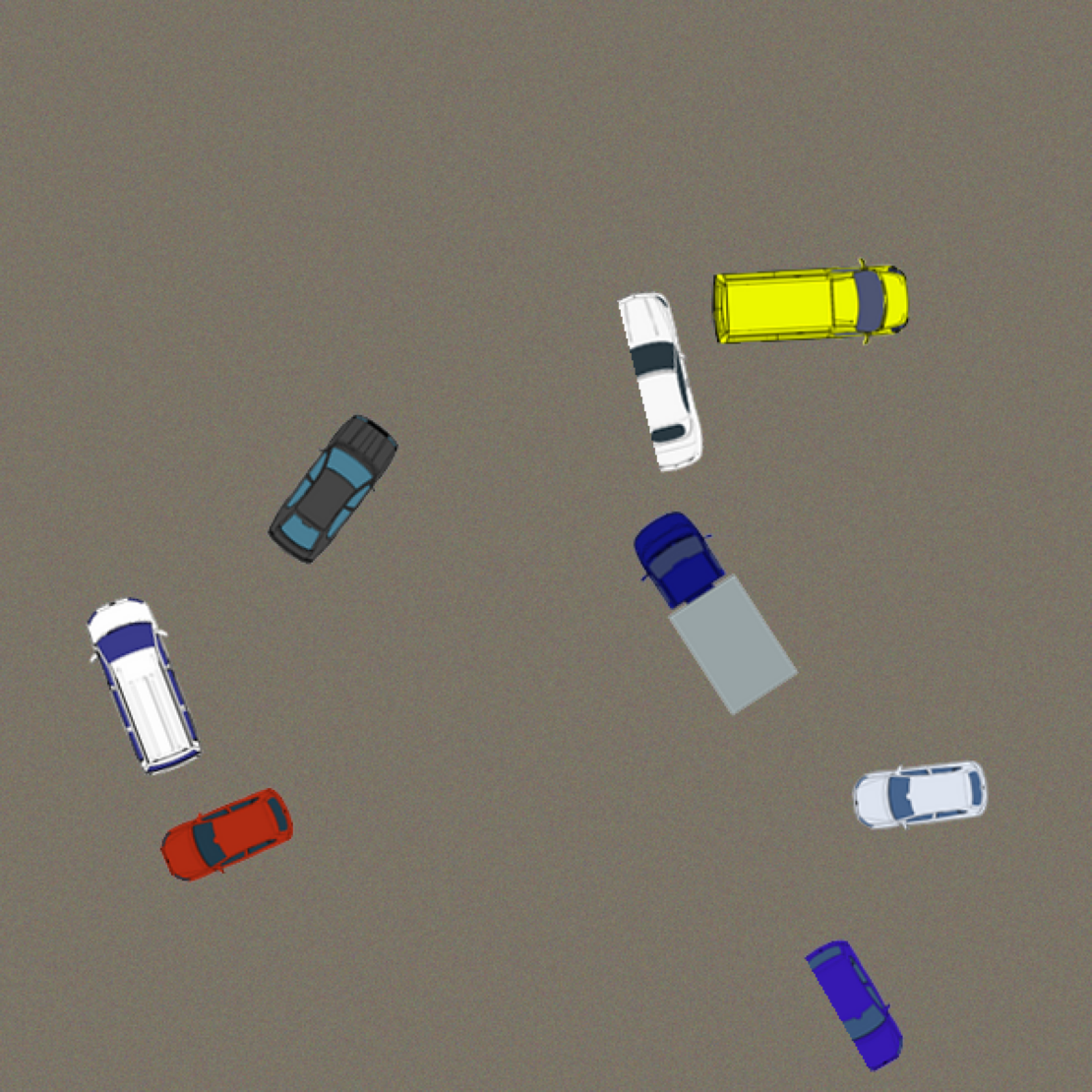}
    \caption{\raggedright + fine background noise} 
    \label{fig:art-gen-pipeline_fine_noise}
    \end{subfigure}
    \hfill
    \begin{subfigure}[t]{0.32\linewidth}
    \includegraphics[width=\linewidth]{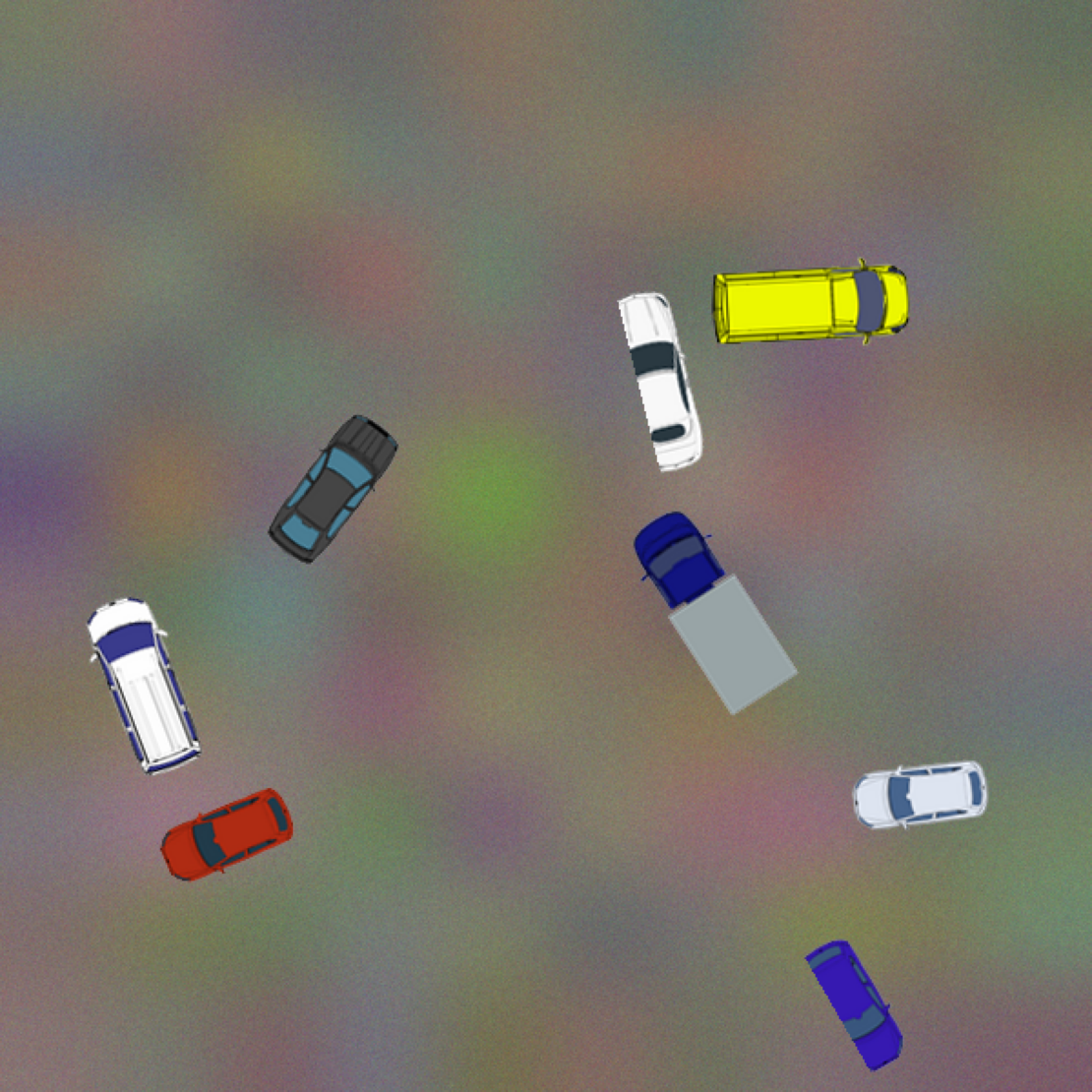}
    \caption{\raggedright + rough background noise} 
    \label{fig:art-gen-pipeline_rough_noise}
    \end{subfigure}
    
    \medskip
    
    \begin{subfigure}[t]{0.32\linewidth}
    \includegraphics[width=\linewidth]{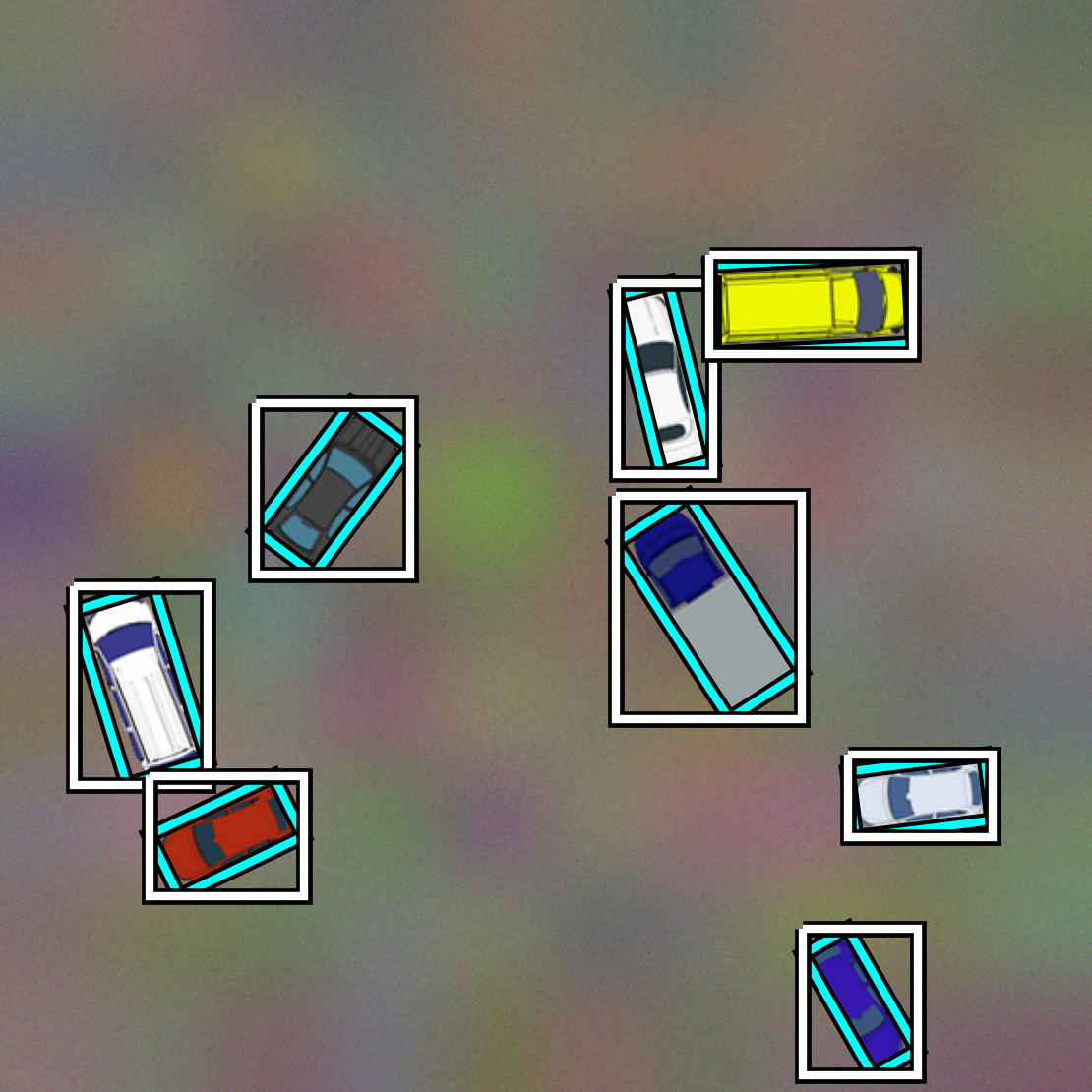}
    \caption{\raggedright + annotations: object and axis aligned bounding boxes} 
    \label{fig:art-gen-pipeline_annotation}
    \end{subfigure}
    \hfill
    \begin{subfigure}[t]{0.32\linewidth}
    \includegraphics[width=\linewidth]{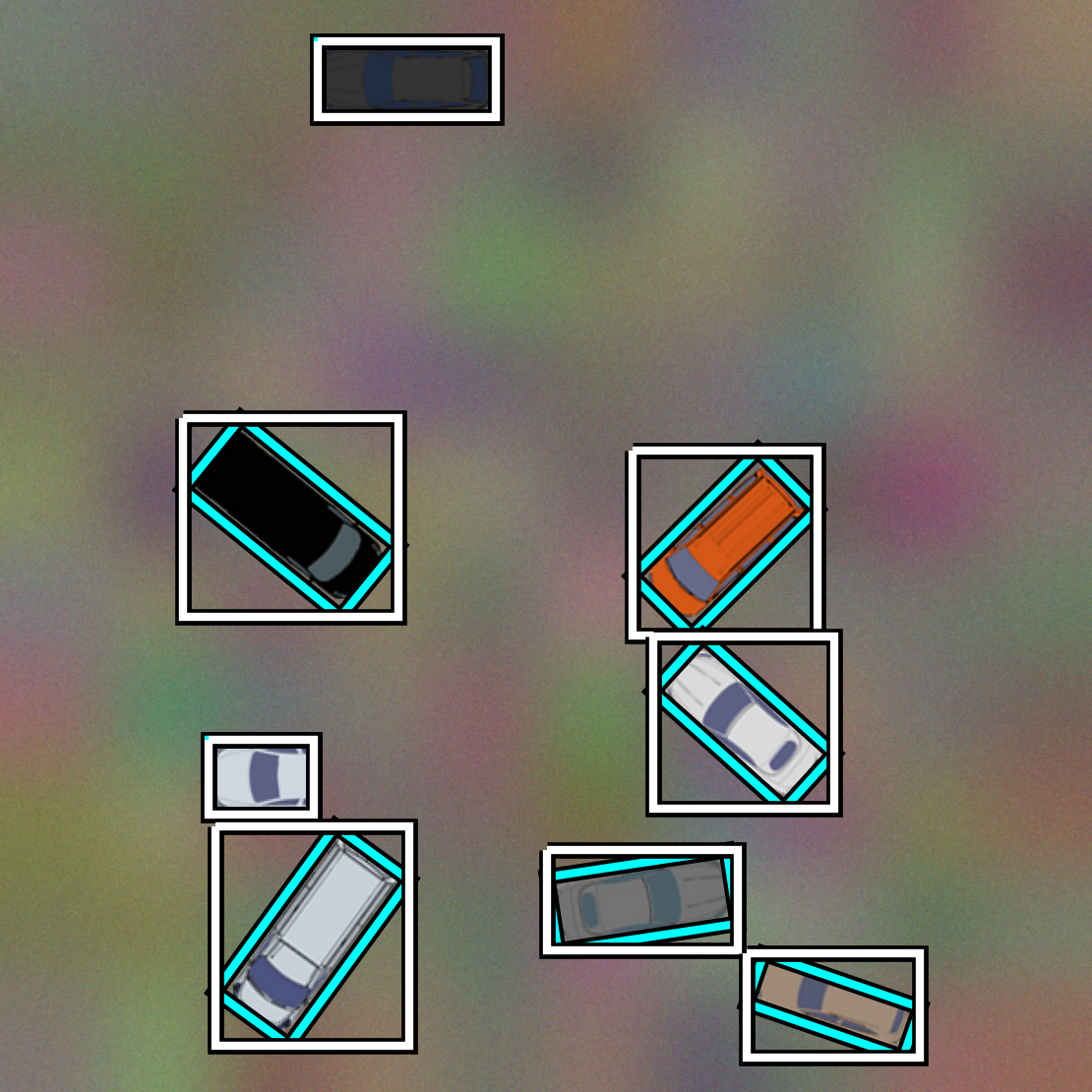}
    \caption{\raggedright additional example} 
    \label{fig:art-gen-pipeline_annotation-sample1}
    \end{subfigure}
    \hfill
    \begin{subfigure}[t]{0.32\linewidth}
    \includegraphics[width=\linewidth]{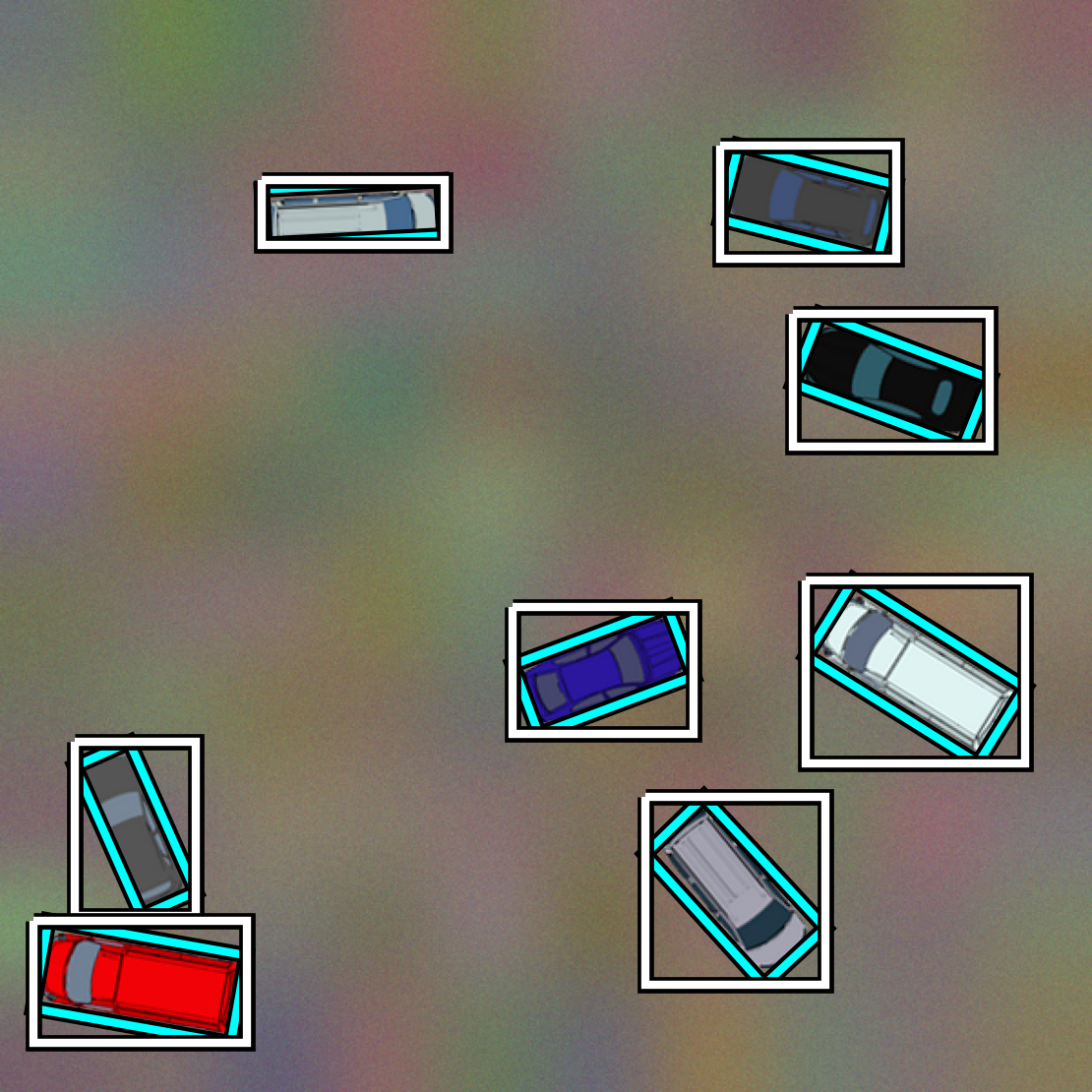}
    \caption{\raggedright additional example} 
    \label{fig:art-gen-pipeline_annotation-sample2}
    \end{subfigure}
    
    \caption{Image generation steps -- Beginning with plain colored vehicles, black outlines, partially depicted vehicles, deformations and noise are added. Annotations for axis- (cyan) and object-aligned (white) bounding boxes are derived automatically.}
    \label{fig:art-gen-pipeline}
\end{figure}

Since the selection of vehicles only corresponds to a tiny subset of the real distribution, we artificially increase the instances' variation by a deformation up to $\pm$~5~\% per side.
The creation of a vehicle is shown in Figure \ref{fig:art-gen-instancing}. 
Each of the instances is randomly rotated and placed on the background canvas while ensuring they do not overlap.
An overview of these steps is given in Figure \ref{fig:art-gen-pipeline}.
Besides measuring the overall effect of the images in Section \ref{sec_sub:art_and_real}, we will investigate the details of this pipeline, like the $\sigma^2$ of the rough noise and the deformation value, in Section \ref{sec_sub:art-ablation}.

\section{Data}
\label{sec:data}

We investigate the presented topic by simulating the lack of data based on an existing large dataset.
This removes the burden of creating a suitable dataset and allows us to determine baseline performance measurements.
Multiple datasets contain imagery and annotations suitable for object and especially vehicle detection in top-down aerial images, which have been reviewed by Azimi et al.~\cite{majid2020eagle} and Cazzato et al.~\cite{cazzato2020survey}.
Datasets like DOTA~\cite{xia2018dota}, COWC~\cite{mundhenk2016large}, VEDAI~\cite{razakarivony2016vehicle}, DLR-MVDA~\cite{liu2015fast}, DLR-Skyscapes~\cite{azimi2019skyscapes} and recently EAGLE~\cite{majid2020eagle}, vary in resolution, type and quality of annotation, numbers of objects, classes and images.

\begin{figure}
    \centering
    
    \begin{subfigure}[t]{0.24\linewidth}
    \includegraphics[width=\linewidth]{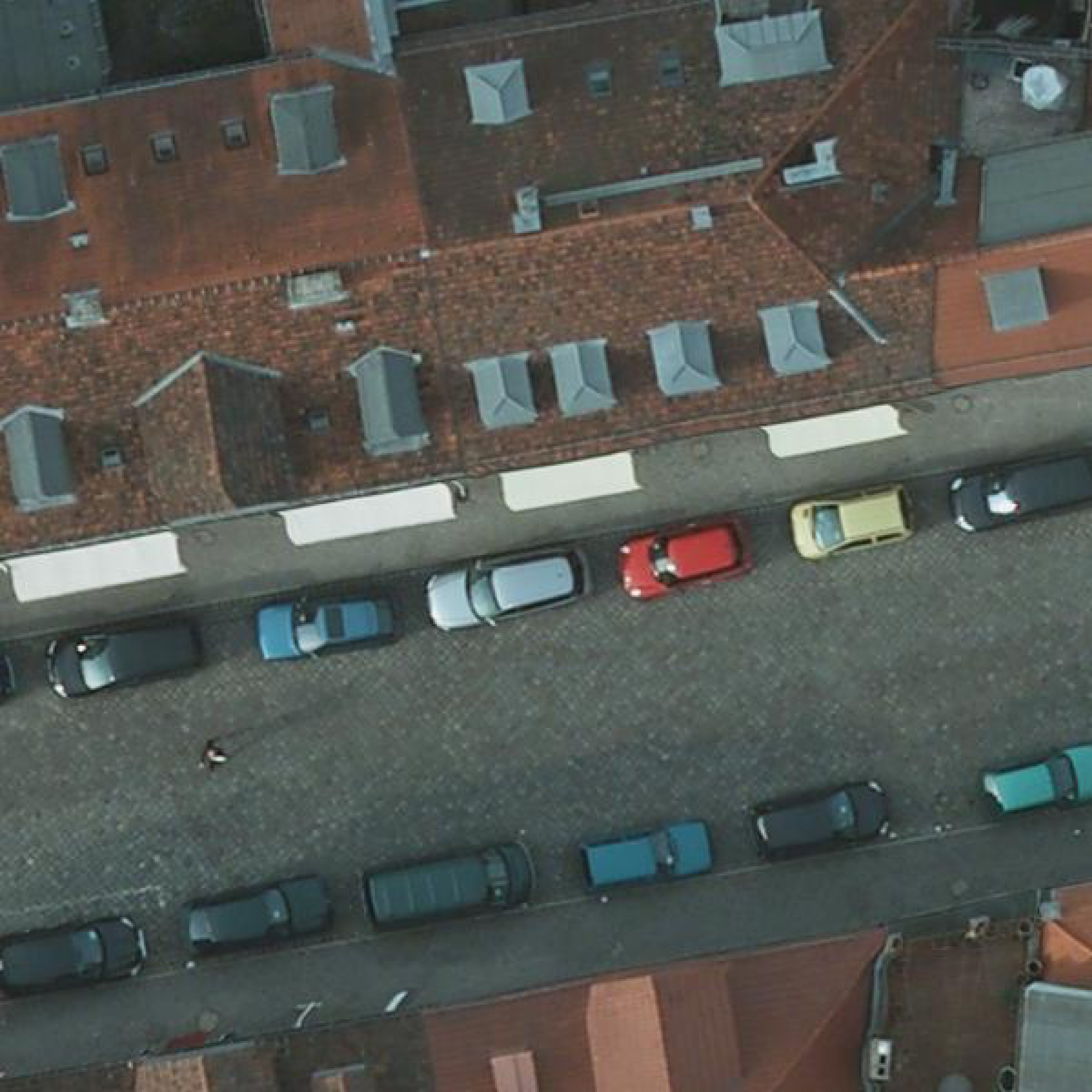}
    \caption{\raggedright RGB image} \label{fig:data-real_trafo_rgb}
    \end{subfigure}
    \hfill
    \begin{subfigure}[t]{0.24\linewidth}
    \includegraphics[width=\linewidth]{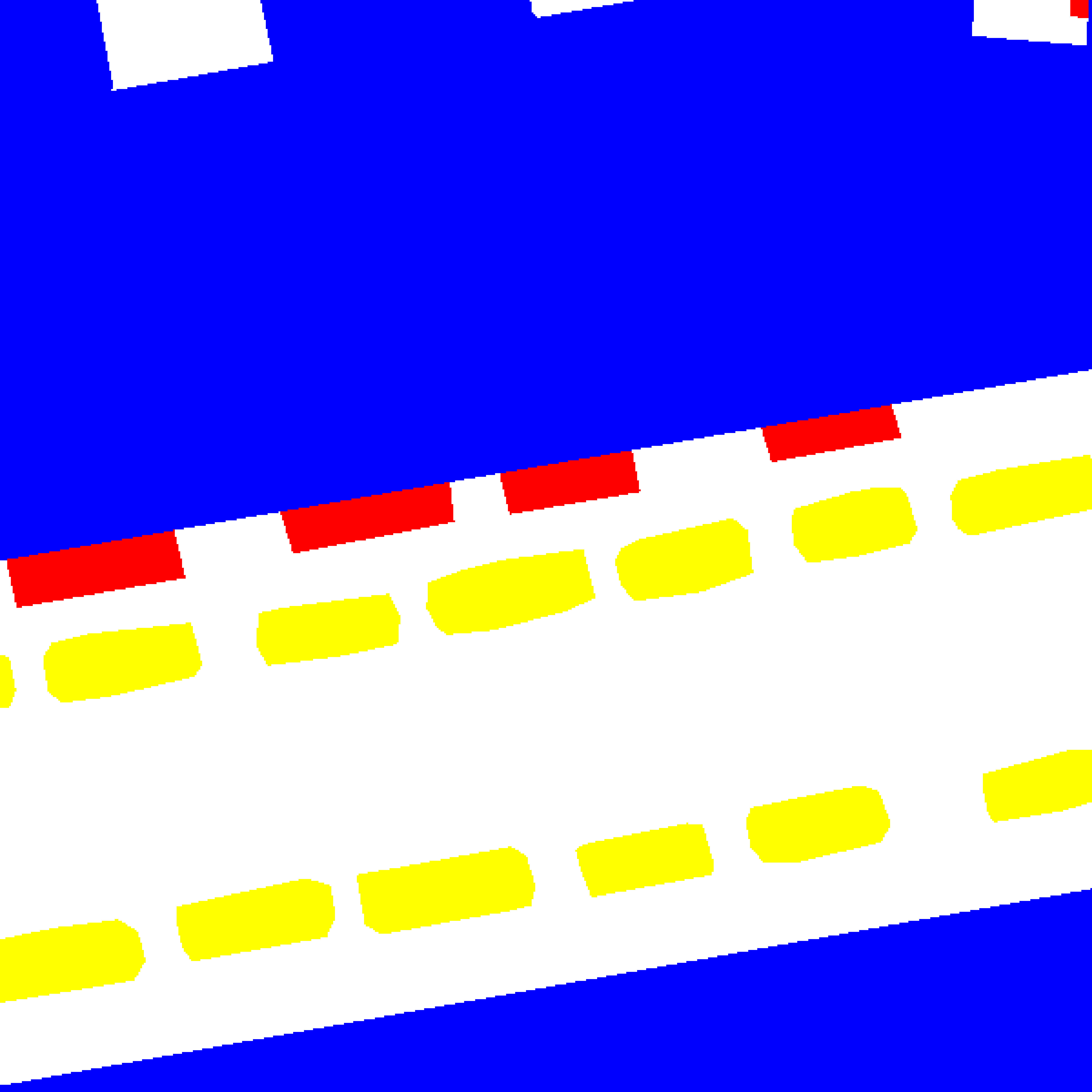}
    \caption{\raggedright semantic annotation} \label{fig:data-real_trafo_label}
    \end{subfigure}
    \hfill
    \begin{subfigure}[t]{0.24\linewidth}
    \includegraphics[width=\linewidth]{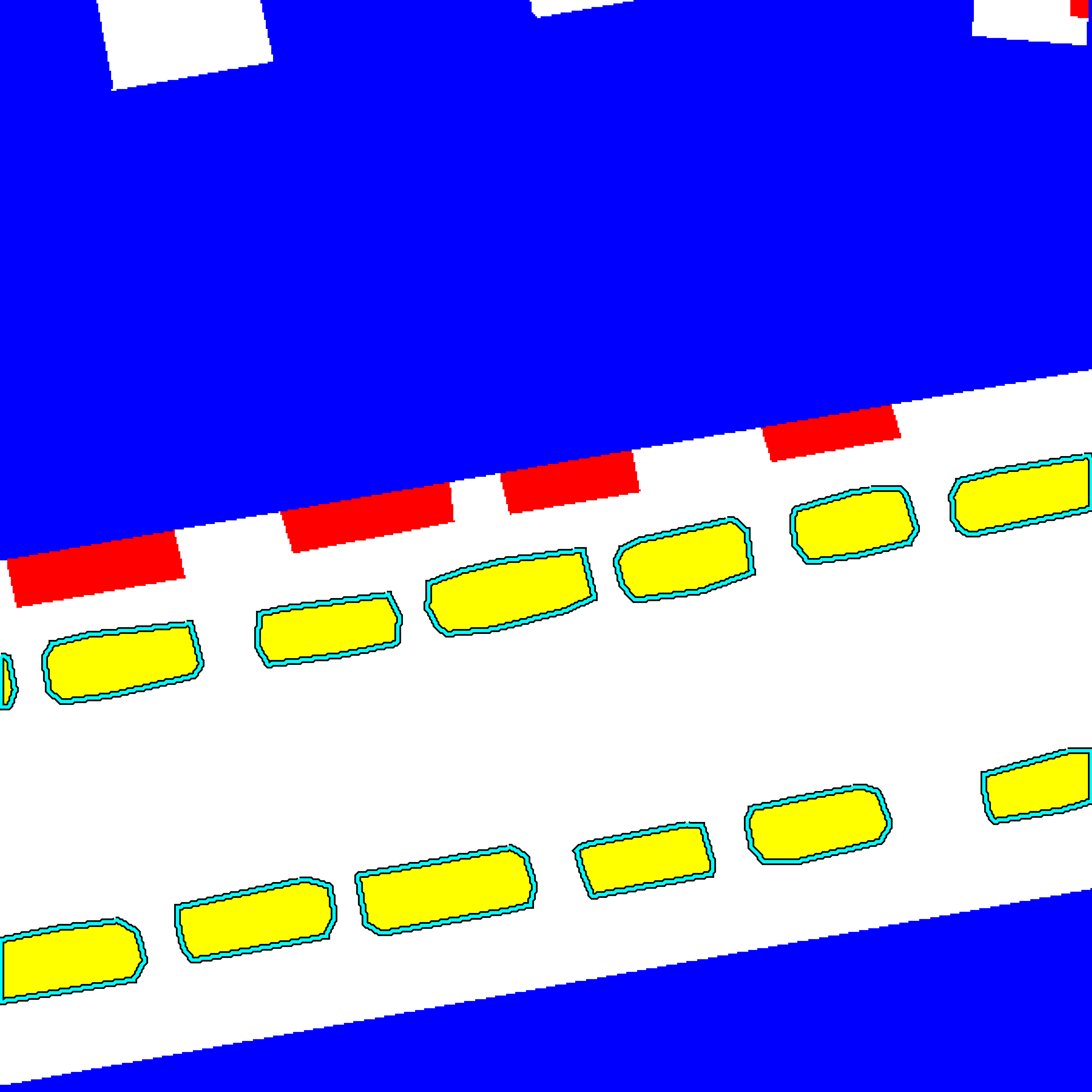}
    \caption{\raggedright object contours (cyan)} \label{fig:data-real_trafo_contours}
    \end{subfigure}
    \hfill
    \begin{subfigure}[t]{0.24\linewidth}
    \includegraphics[width=\linewidth]{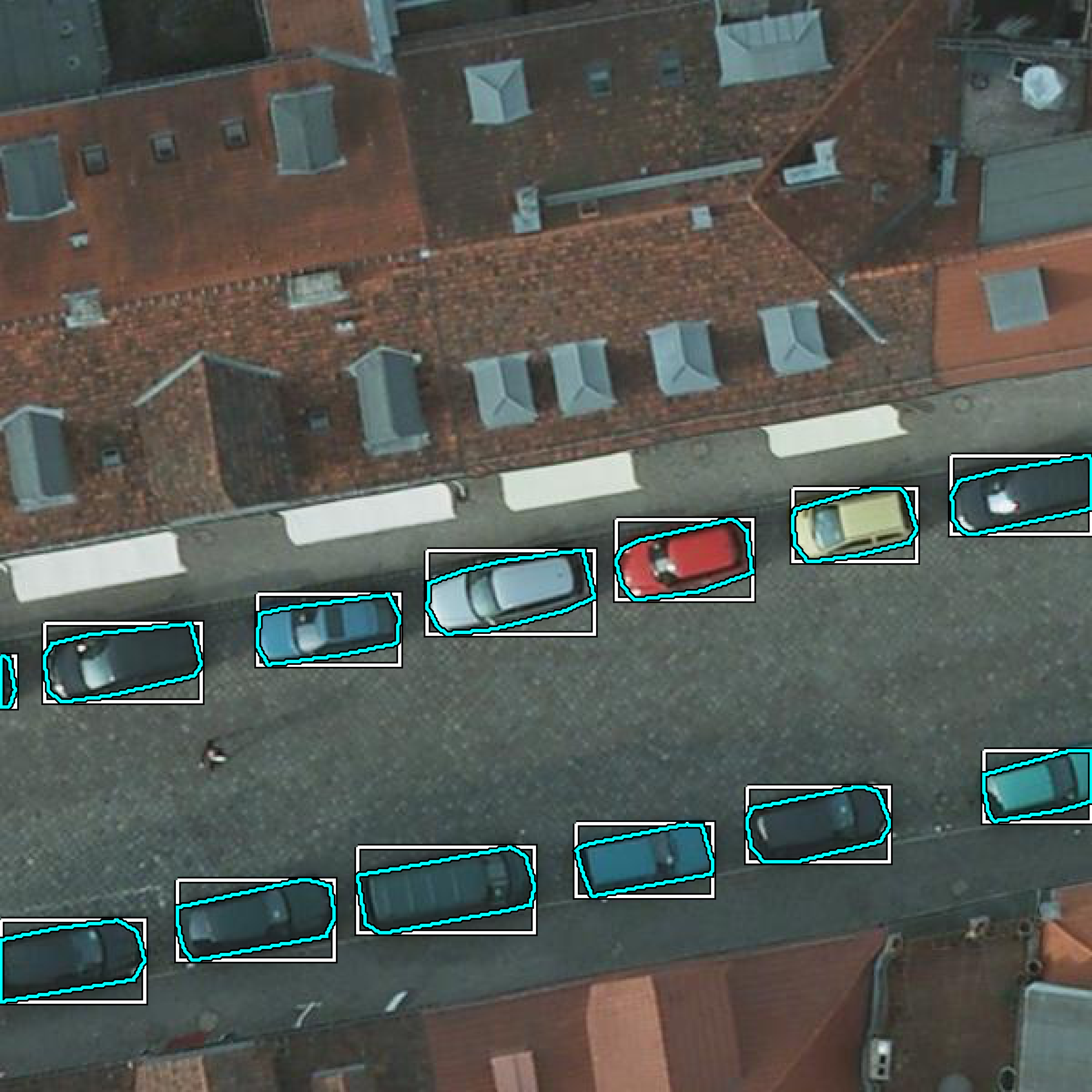}
    \caption{\raggedright derived bounding boxes} \label{fig:data-real_trafo_minarearect}
    \end{subfigure}
\caption{Transformation of the semantic annotation to bounding boxes -- The semantic image (\subref{fig:data-real_trafo_label}) is used to find contours of yellow image areas, that are annotated to be class \texttt{car} (\subref{fig:data-real_trafo_rgb}), which are used to derive the cyan object- and white axis-aligned bounding boxes (\subref{fig:data-real_trafo_minarearect}).}
\label{fig:data-real_trafo}
\end{figure}

Considering suitability and availability, we use a modified version of the Potsdam dataset\footnote{http://www2.isprs.org/commissions/comm3/wg4/2d-sem-label-potsdam.html}, which is part of the ISPRS 2D Semantic Labeling Contest~\cite{rottensteiner2013}.
As we do not intend to provide new high scores but work on other aspects, we decided on this large-scale dataset since it matches our actual domain the best, and we modify it to fit our requirements.
The Potsdam dataset is created from a true ortho-mosaic with a ground sampling distance of 0.05 \si{\meter/\px} of the city of Potsdam and contains thousands of vehicles.
The mosaic has been cut into 38 image tiles of the size 6000 $\times$ 6000 \si{\px}, divided into 24 tiles for training and 14 tiles for testing.
The dataset contains pixel-level semantic annotations of the six classes \texttt{impervious surfaces}, \texttt{building}, \texttt{low vegetation}, \texttt{tree}, \texttt{car}, and \texttt{clutter background}, which are color-coded by distinct colors.
As we concentrate on vehicle detection, we only use the class \texttt{car}, which contains cars themselves, vans, and some trucks.
We manually remove a single tram that is also annotated.
Being originally designed for semantic segmentation, we modify the dataset to fit our needs best.
We use image processing to find the contours of the image areas annotated with the class-specific color and then derive object-aligned and axis-aligned bounding boxes for each contour.
A sample of this process is shown in Figure \ref{fig:data-real_trafo}.

As already pointed out in \cite{weber2020learning} there are erroneous annotations and artifacts caused by the structure-from-motion process, which was used to create the true ortho-mosaic.
Some connected instances have been manually split, and the artifacts might have a small impact on detection quality.
However, the biggest issue for object detection is the annotation policy, where the structurally top-most object decides the class.
The dataset contains multiple vehicles partially or entirely located under trees, and hence, these vehicles are only partially or not annotated.
Since the images were acquired in autumn or winter, many of the trees are leave-less, and the annotation practice becomes an issue.
The majority of these vehicles is visible to the human eye and the object detector, leading to, formally, false-positive detections, which degrade the performance.
These missing annotations, whose impact has been analyzed by Xu et al.\cite{xu2019missing}, are partially to blame for a remaining performance gap.

\section{Experiments}
\label{sec:experiments}

In the following subsections, we will first describe the setup of our experiments in Subsection \ref{sec_sub:experimental_setup}.
In Subsection \ref{sec_sub:art_and_real} we continue by showing the positive effect of our artificial images when only a few real images are available by comparing models trained with a varying amount of real images and models trained with additional artificial images.
In the same subsection, we show that our approach is agnostic towards the capacity of the feature extraction network, the ground sampling distance, and that image augmentation is a complementary approach.
To provide a better understanding of the remaining performance gap, we will go through the components of the image generator in Subsection \ref{sec_sub:art-ablation}. 
We close the experiments by showing an analysis of the effect of the background and image composition in Subsection \ref{sec_sub:art-composition}. 

\subsection{Experimental Setup}
\label{sec_sub:experimental_setup}

For the following experiments, we use the Potsdam dataset imagery described in Section \ref{sec:data}.
Since the Potsdam images are too large to be processed as a whole, we cut them into smaller patches suitable for our object detector.
These patches' size controls the derived dataset's size, as the area covered by the Potsdam images is fixed.
Smaller patches result in larger datasets and allow for larger batch sizes and vice versa.
We further can artificially enlarge the ground sampling distance (GSD) of the data by down-sampling the patches.
This gives the option to reduce the dataset size and training times while keeping smaller images' benefits.
In advance of our experiments, we evaluated the interplay of the patch size, GSD, and dataset size.
The evaluation revealed that small input sizes combined with large batch sizes and large dataset sizes lead to better model training behavior.
However, too large datasets result in prolonged training times while giving only minor performance improvements.
Based on these findings, we identified the following combination as a reasonable compromise between model performance and training times.
We create patches of size 600 $\times$ 600 \si{\px} and down-sample them to the object detector's input size of 300 $\times$ 300 \si{\px}, which increases the original GSD of 0.05~\si{\meter/\px} to 0.10~\si{\meter/\px}.
In Section \ref{sec_sub:art_and_real-gsd} we address this GSD enlargement and show that the resulting loss in performance can be neglected.
The resulting dataset consists of 5400 patches with an overlap of 200~\si{\px} for the training section and 3150 patches for the test section with 12842 and 3859 vehicles.

In the dataset, some vehicles are partially concealed by buildings or trees or cut by the patching process.
We remove annotations, which are affected by this and are smaller than 20 \si{\px}.
Afterward, we remove images that do not contain any vehicle.
We separate 30 \% of the training images to create a validation dataset.
This results in 2039 images for training, 907 images for validation, and 834 for testing, with 8980, 3862, and 3859 vehicles, respectively.

Our artificial images are based on eight blueprints of cars.
They consist of compact cars, sedans and station wagons of mid and top-range cars, small and large vans, and small transporters.
These blueprints cover the common vehicles in the Potsdam dataset.
Each image contains ten vehicles, of which three are only partial vehicles.
A small test revealed that the performance is not sensitive to the number of artificial images within a wide range.
However, with 1000 artificial images, the best detection performance can be achieved.

Our code used for the experiments is implemented with the pytorch~\cite{paszke2019pytorch} framework and the high-level training library pytorch-lightning~\cite{falcon2019pytorch}.
We use a pre-trained feature extraction network for our detector, which by default is a ResNet-50.
It has been trained with the ImageNet~\cite{deng2009imagenet} dataset and is part of the torchvision model zoo \cite{paszke2019pytorch}.
Our data is standard-normalized with the ImageNet mean color. 
We use random flipping of the x- and y-axis and random changes of brightness, contrast, hue, and saturation as our default augmentations.
If possible, we use a batch size of 32 given the dataset size, or 16 or 8, respectively.
Larger batch sizes do not show an improvement in any aspect.
We optimize our models using SGD and alter learning rate and momentum using the one-cycle policy \cite{smith2018disciplined}.
It smoothly ramps up the learning from 1/25th of the peak learning rate of 0.0075, which is reached after 30 \% of the training, and afterward lowered to 1/10000th of it.
Simultaneously, the momentum starts with 0.95, is lowered to 0.85, and raised again to 0.95.
We choose to train each model with a fixed number of steps, 2500 if not specified otherwise, rather than epochs.
This allows us to have a constant number of iterations regardless of the dataset size.
We keep track of the current best model and load it afterward to get the best model in case of over-fitting and divergence.
The training process is stopped early when the validation loss does not decrease within five epochs.
Training is done on single NVIDIA Tesla V100 GPUs with 32 \si{\giga\byte} memory.
It typically takes 15 minutes and consumes six \si{\giga\byte} of GPU memory in mixed precision memory mode.

For evaluation, we filter the predicted bounding boxes by their confidence score with a threshold of 0.1.
We compute the average precision (AP), as described for the MS COCO dataset \cite{lin2014microsoft}.
Our intersection-over-union (IoU) threshold is 0.5, which matches MS COCO's AP\textsuperscript{IoU=0.5}.
All experiments are repeated eight times, and we report mean and standard deviations for the average precision, denoted by $\mathrm{AP}_{\mathrm{\mu}}$ and $\mathrm{AP}_{\mathrm{\sigma}}$, respectively.
We mark them with $\uparrow$ for higher is better and with $\downarrow$ for the opposite.

\subsection{Combining real and artificial images}
\label{sec_sub:art_and_real}

\begin{figure}
  \centering

  \includegraphics[width=\linewidth]{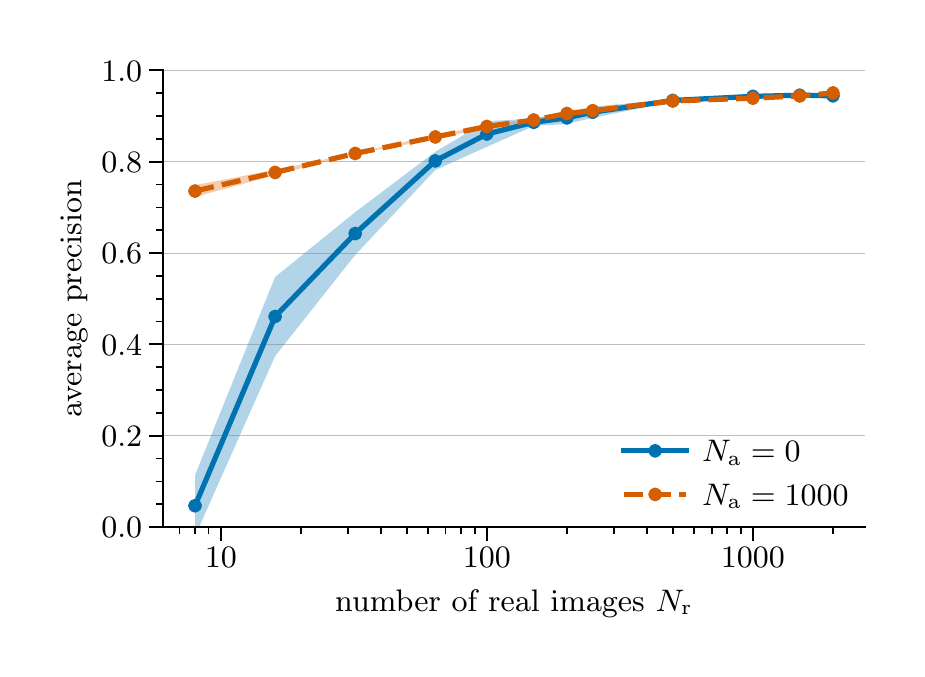}
  
  \caption{Performance comparison for models trained with baseline (blue) and combined datasets (orange) -- The x-axis shows the number of real images ($N_{\mathrm{r}}$) on a logarithmic scale. The positive effect of the 1000 artificial images ($N_{\mathrm{a}}$) is most significant for the numbers of real images below 100, whereas it diminishes for larger numbers.}
  \label{fig:art_and_real}
\end{figure}

In this first experiment, we show that our artificial images help when only small amounts of real images are available. 
For this, we create two groups of datasets:
\begin{itemize}
    \item baseline: 12 datasets with 8 to 2039 real images, and
    \item combination: 12 datasets with 8 to 2039 real images and 1000 artificial images.
\end{itemize}
We train a model for each of these datasets and compare these two groups in Figure \ref{fig:art_and_real}, where we show the average precision of these models.

The solid blue curve shows the baseline models' performance.
The average precision with datasets consisting only of a few images is relatively low.
With increased dataset size, the average precision rises quickly and passes 0.8 at approximately 100 images and 0.9 at 200 images.
With larger numbers of real images, the average precision converges, reaching 0.95 for the largest dataset with 2039 real images.

The dashed orange curve shows the average precision of the second group models, which are trained with the combination of real and artificial images.
All of these models have an average precision larger than 0.7.
The average precision levels of 0.8 and 0.9 are reached with 32, respectively, 100 real images.
These numbers are about half of the real images needed for the same levels when no artificial images are used.
With larger dataset sizes, the effect of adding artificial images reduces.
For the largest dataset with 2039 real and 1000 artificial images, the average precision converges as well to 0.95.

\begin{figure}
    \centering
    
    \begin{subfigure}[t]{0.24\linewidth}
    \includegraphics[width=\linewidth]{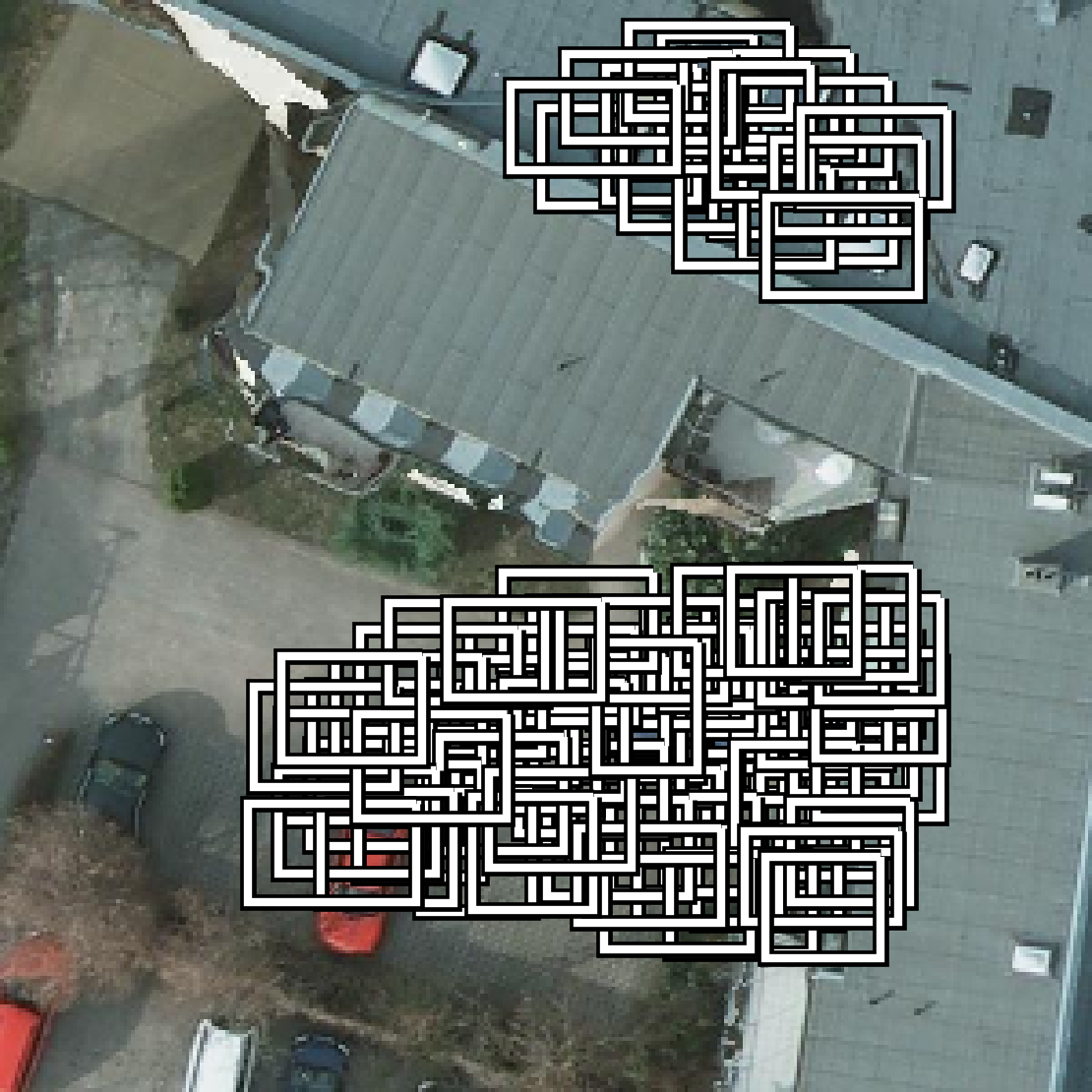}
    \caption{\raggedright $N_{\mathrm{r}}=8, N_{\mathrm{a}}=0$} 
    \label{fig:art_and_real-samples-8-0}
    \end{subfigure}
    \hfill
    \begin{subfigure}[t]{0.24\linewidth}
    \includegraphics[width=\linewidth]{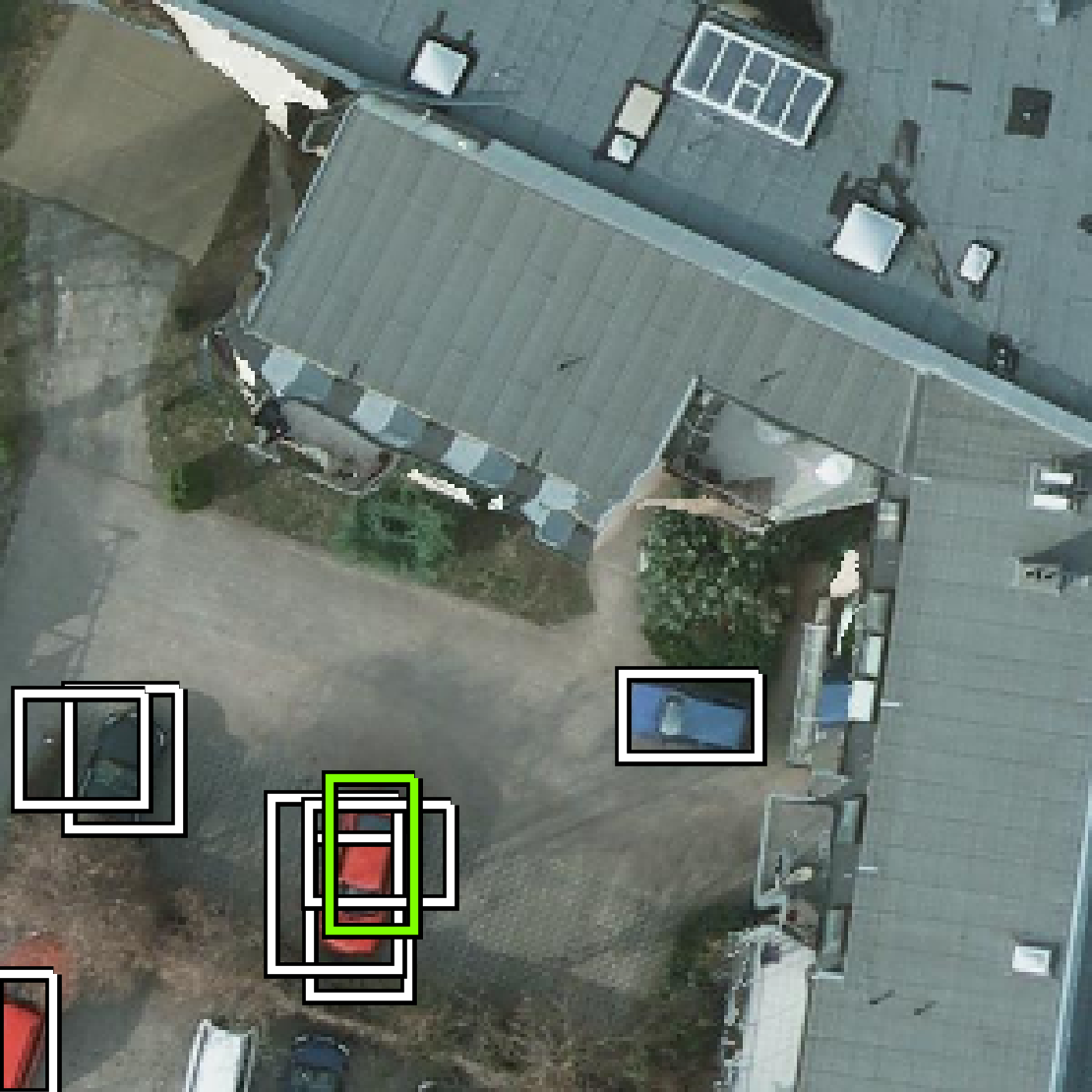}
    \caption{\raggedright $N_{\mathrm{r}}=32, N_{\mathrm{a}}=0$} 
    \label{fig:art_and_real-samples-32-0}
    \end{subfigure}
    \hfill
    \begin{subfigure}[t]{0.24\linewidth}
    \includegraphics[width=\linewidth]{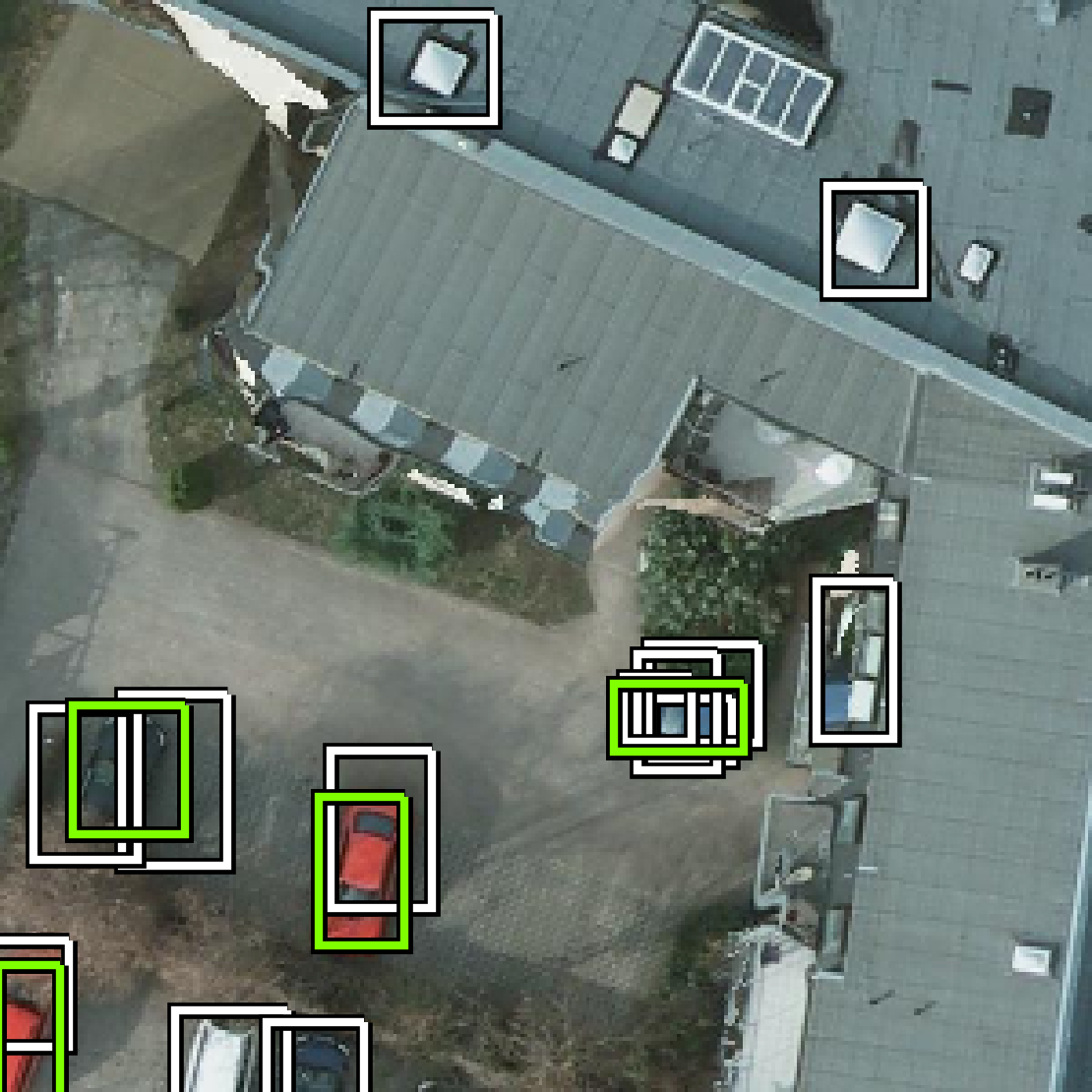}
    \caption{\raggedright $N_{\mathrm{r}}=64, N_{\mathrm{a}}=0$}
    \label{fig:art_and_real-samples-64-0}
    \end{subfigure}
    \hfill
    \begin{subfigure}[t]{0.24\linewidth}
    \includegraphics[width=\linewidth]{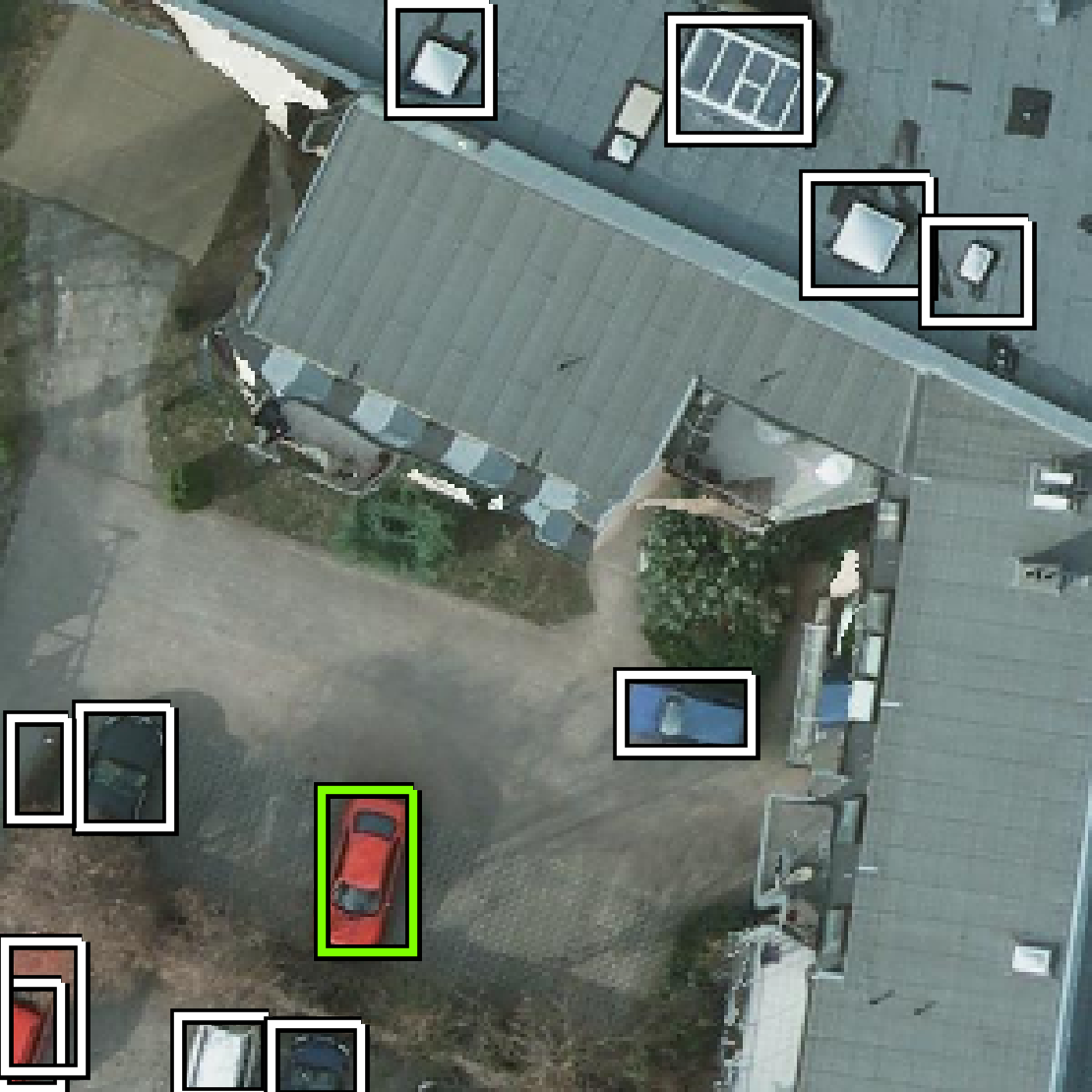}
    \caption{\raggedright $N_{\mathrm{r}}=150, N_{\mathrm{a}}=0$}
    \label{fig:art_and_real-samples-150-0}
    \end{subfigure}
    
    \medskip
    
    \begin{subfigure}[t]{0.24\linewidth}
    \includegraphics[width=\linewidth]{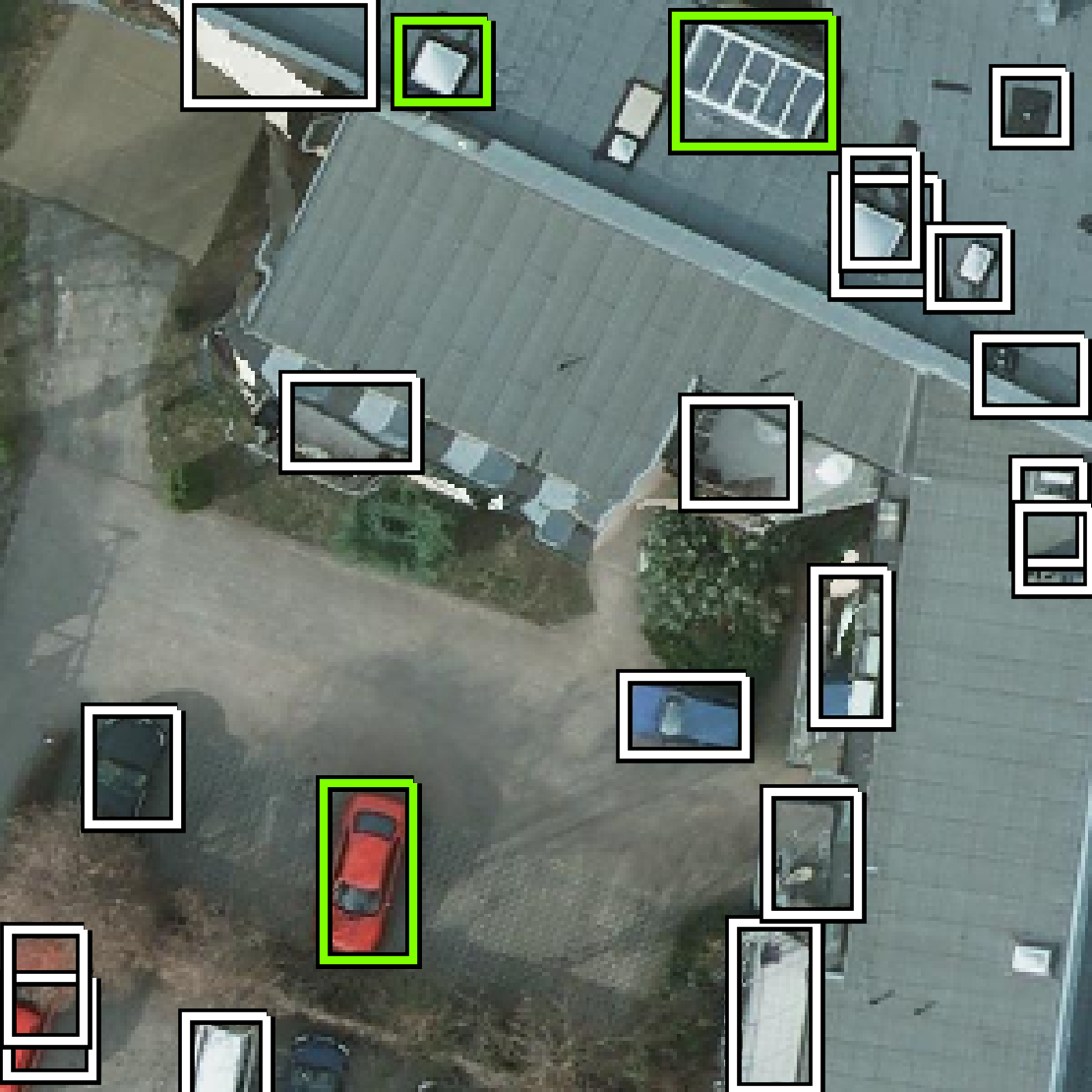}
    \caption{\raggedright $N_{\mathrm{r}}=8, N_{\mathrm{a}}=1000$} 
    \label{fig:art_and_real-samples-8-1000}
    \end{subfigure}
    \hfill
    \begin{subfigure}[t]{0.24\linewidth}
    \includegraphics[width=\linewidth]{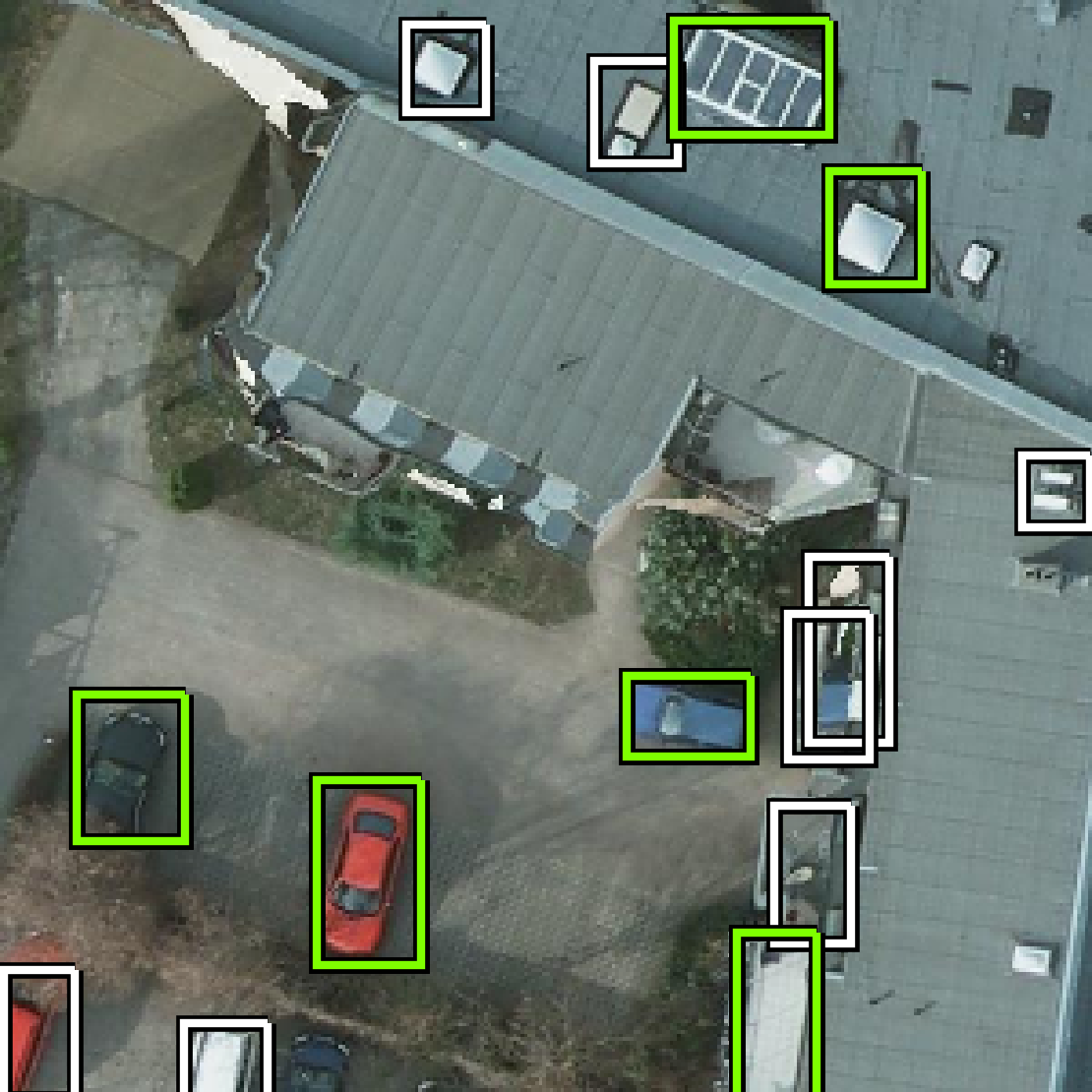}
    \caption{\raggedright $N_{\mathrm{r}}=32, N_{\mathrm{a}}=1000$} 
    \label{fig:art_and_real-samples-32-1000}
    \end{subfigure}
    \hfill
    \begin{subfigure}[t]{0.24\linewidth}
    \includegraphics[width=\linewidth]{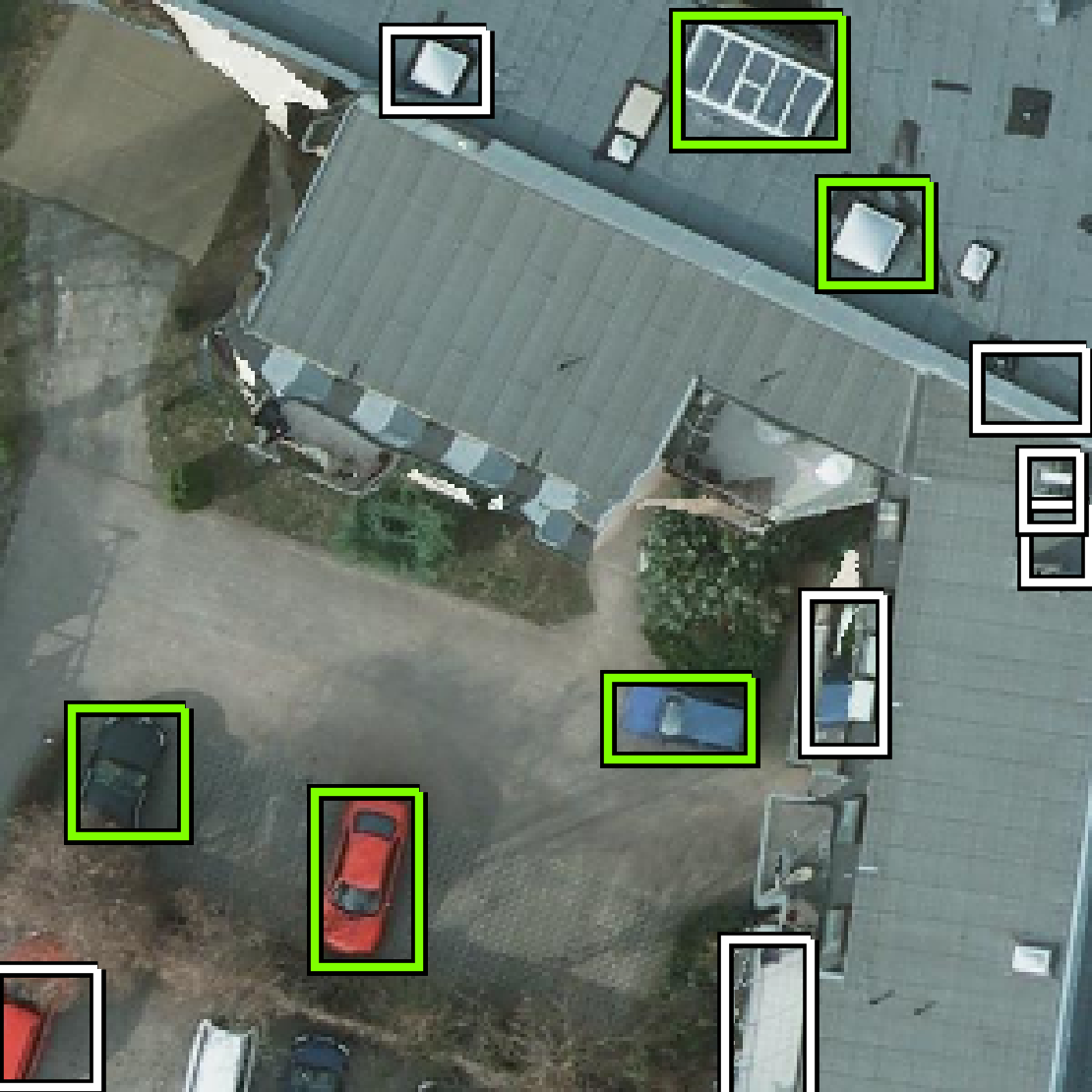}
    \caption{\raggedright $N_{\mathrm{r}}=64, N_{\mathrm{a}}=1000$}
    \label{fig:art_and_real-samples-64-1000}
    \end{subfigure}
    \hfill
    \begin{subfigure}[t]{0.24\linewidth}
    \includegraphics[width=\linewidth]{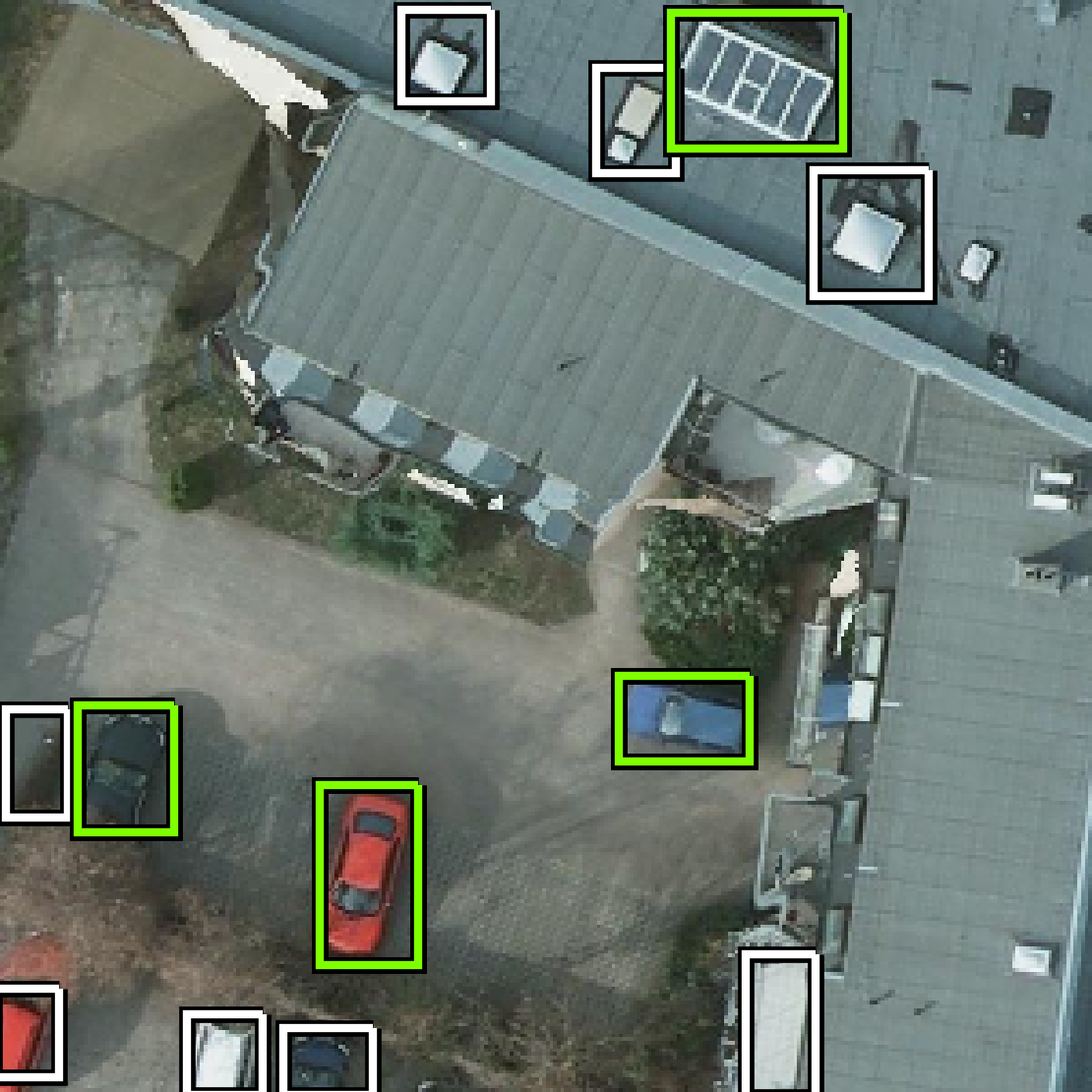}
    \caption{\raggedright  $N_{\mathrm{r}}=150, N_{\mathrm{a}}=1000$} 
    \label{fig:art_and_real-samples-150-1000}
    \end{subfigure}
    
    \caption{Predictions of baseline models (top) and models trained with combined datasets (bottom) -- The baseline models show an overall poor and fluctuating performance. Many vehicles are undetected, multiple predictions per vehicle occur, and since only a few boxes are green, the confidences are low. The second group's models show a stabler result with more green predictions with larger confidences and sharper bounding boxes. $N_{\mathrm{r}}$: number of real images,  $N_{\mathrm{a}}$: number of artificial images; white: confidence $>$ 0.1, green: confidence $>$ 0.5.}
    \label{fig:art_and_real-samples}
\end{figure}

In the following, we make a qualitative comparison of predictions of models from the two dataset groups.
For this purpose, we select the models of the datasets containing 8, 32, 64, and 150 real images since these are in the range of the most notable differences between the two groups (see Fig. \ref{fig:art_and_real}).
Figure \ref{fig:art_and_real-samples} shows the predictions of a single image of the baseline models in the upper row and those of the models trained with the combined datasets in the lower row.
We show bounding boxes in white, which have confidence greater than 0.1 and are thus considered in the evaluation. 
Furthermore, we show bounding boxes, with a confidence greater than 0.5 in green, which gives an impression about confidence levels.
If we look at the upper row, the model of the smallest dataset (Fig. \ref{fig:art_and_real-samples-8-0}) stands out since it does not provide useful predictions.
The remaining models (see Fig. \ref{fig:art_and_real-samples-32-0} to \ref{fig:art_and_real-samples-150-0}), which were trained with an increased number of real images, show a higher detection rate, but false-negative and false-positive detections, and fluctuations in the predicted boxes and confidences, visible in their color changes.
Overall, these models fail to produce predictions with a reliable large confidence value and, in some cases, confuse rectangular image features with vehicles.
Compared to that, the models in the bottom row (see Fig. \ref{fig:art_and_real-samples-8-1000} to \ref{fig:art_and_real-samples-150-1000}) that have been trained with additional artificial images show more stable results.
Whereas the models whose results are shown in the top row often produce multiple predictions per vehicle, the bottom models tend to provide a single bounding box.
Besides, more predictions have larger confidence, illustrated with green bounding boxes.
Although the models learned on the combined datasets also tend to confuse some rectangular image regions for vehicles, their overall quality is better than the upper row models.

In summary, we see that when only a few real images are available, a single image's performance impact is large.
In contrast, once a certain threshold is passed, the impact reduces.
Adding artificial images to a small number of real images largely improves the performance and does not harm if larger amounts are available.

Despite this improvement, there remains a gap between what can be achieved with large numbers of real images compared to artificial images.
With 1000 artificial images and eight real images, we achieve 0.70~AP, whereas 1000 real images result in 0.94~AP.
It appears that the usable information content of a single artificial image is lower than that of a real image.
We provide further investigation into this in Sections \ref{sec_sub:art-ablation} and \ref{sec_sub:art-composition}.
Before that, we verify the tolerance to network capacity, ground sampling distance, and image augmentation in the following paragraphs.

\paragraph{Impact of network capacity}
\label{sec_sub:art_and_real-capacity}

In this section, we evaluate the influence of the feature extraction network's capacity.
For this, we select a shallow ResNet-18, a medium ResNet-50 and a deep ResNet-152 \cite{he2016deep}.
In addition, we use a more recent ResNeXt-50 \cite{xie2017aggregated} and a WideResNet-50 \cite{zagoruyko2016wide}.
Table \ref{tab:art_and_real-capacity} shows the detection performance of these models trained with five different datasets.
These include our baseline dataset, containing 2039 real images, and datasets with 8 and 150 real images, and each of them combined with 1000 artificial images.
Each column in the table is labeled with the number real~($N_{\mathrm{r}}$) and artificial images~($N_{\mathrm{a}}$).

\begin{table*}
\small
\centering
\caption{Object detection performance of baseline models and models trained with combined images for different feature extraction networks -- 
Most models share similar performance levels with our default architecture, ResNet-50. ResNet-18 shows a slightly worse performance on the complete real-world data set, and it also benefits less from additional artificial images. $N_{\mathrm{r}}$: number of real images,  $N_{\mathrm{a}}$: number of artificial images; $(\cdot)$*: default choice.}
\label{tab:art_and_real-capacity}
\begin{tabular}{c|cc|cccc|cccc}
\toprule
$N_{\mathrm{r}} + N_{\mathrm{a}}$ & \multicolumn{2}{c|}{2039 + 0} & \multicolumn{2}{c}{8 + 0} & \multicolumn{2}{c|}{8 + 1000} & \multicolumn{2}{c}{150 + 0} & \multicolumn{2}{c}{150 + 1000} \\
{} & AP$_{\mathrm{\mu}}\uparrow$ & AP$_{\mathrm{\sigma}}\downarrow$ & AP$_{\mathrm{\mu}}\uparrow$ & AP$_{\mathrm{\sigma}}\downarrow$ & AP$_{\mathrm{\mu}}\uparrow$ & AP$_{\mathrm{\sigma}}\downarrow$ & AP$_{\mathrm{\mu}}\uparrow$ & AP$_{\mathrm{\sigma}}\downarrow$ & AP$_{\mathrm{\mu}}\uparrow$ & AP$_{\mathrm{\sigma}}\downarrow$ \\
network       &                             &                                  &                             &                                  &                             &                                  &                             &                                  &                             &                                  \\
\midrule
ResNet-18     &                       0.934 &                            0.015 &                       0.000 &                            0.000 &                       0.665 &                            0.020 &                       0.876 &                            0.005 &                       0.879 &                            0.007 \\
ResNet-50*     &                       0.946 &                            0.008 &                       0.047 &                            0.067 &                       0.740 &                            0.013 &                       0.886 &                            0.008 &                       0.892 &                            0.003 \\
ResNet-152    &                       0.949 &                            0.004 &                       0.025 &                            0.039 &                       0.736 &                            0.017 &                       0.895 &                            0.009 &                       0.893 &                            0.005 \\
ResNeXt-50    &                       0.948 &                            0.003 &                       0.091 &                            0.103 &                       0.739 &                            0.011 &                       0.898 &                            0.008 &                       0.899 &                            0.002 \\
WideResNet-50 &                       0.947 &                            0.003 &                       0.064 &                            0.035 &                       0.742 &                            0.011 &                       0.898 &                            0.007 &                       0.899 &                            0.005 \\
\bottomrule
\end{tabular}
\end{table*}

For the baseline data sets in the second column, marked with 2039 + 0, all networks' performance is on par within 0.02 points of AP.
The results over several runs show a stable behavior as indicated by the standard deviations.
The center models, trained with 8 real images, also have similar gains when 1000 artificial images are added.
All of the models on the right also show the saturation behavior of our default network, ResNet-50.
Overall, the ResNet-18 performs slightly worse and does not gain as much as the networks do.

These results show that artificial images have a positive impact regardless of the feature extraction network's capacity.
However, the best performance requires a network with sufficient capacity, like ResNet-50.
It also shows that larger networks do not perform better, and therefore the associated longer training times, inference times, and larger memory requirements can be avoided.

\paragraph{Impact of ground sampling distance}
\label{sec_sub:art_and_real-gsd}

\begin{figure}
    \centering
    
    \begin{subfigure}[t]{0.24\linewidth}
    \includegraphics[width=\linewidth]{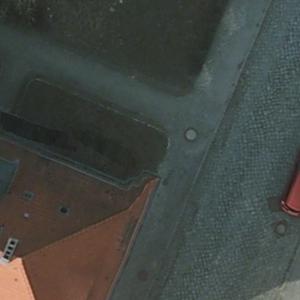}
    \caption{\raggedright GSD 0.05 \si{\meter/\px}} \label{fig:art_and_real-gsd-300}
    \end{subfigure}
    \hfill
    \begin{subfigure}[t]{0.24\linewidth}
    \includegraphics[width=\linewidth]{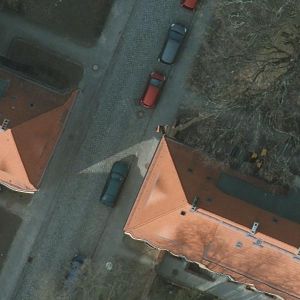}
    \caption{\raggedright GSD 0.10 \si{\meter/\px}} \label{fig:art_and_real-gsd-600}
    \end{subfigure}
    \hfill
    \begin{subfigure}[t]{0.24\linewidth}
    \includegraphics[width=\linewidth]{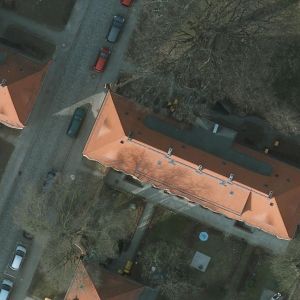}
    \caption{\raggedright GSD 0.15 \si{\meter/\px}} \label{fig:art_and_real-gsd-900}
    \end{subfigure}
    \hfill
    \begin{subfigure}[t]{0.24\linewidth}
    \includegraphics[width=\linewidth]{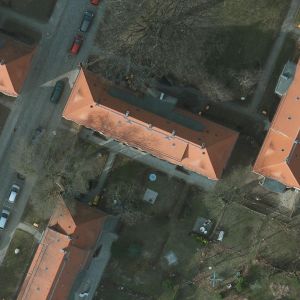}
    \caption{\raggedright GSD 0.2 \si{\meter/\px}} \label{fig:art_and_real-gsd-1200}
    \end{subfigure}
    
    \caption{Samples of the selected GSDs -- The different fields of view and the loss of detail are apparent.}
    \label{fig:art_and_real-gsd}
\end{figure}

The ground sampling distance (GSD) is an essential parameter for the acquisition of aerial images.
We described in Subsection \ref{sec_sub:experimental_setup}, that compared to the original Potsdam imagery we use an increased GSD of 0.01 \si{\meter/\px}.
In order to analyze this in more detail and underline the suitability of our choice, 
we test the applicability of our artificial images with various GSDs, which are: 1) the original GSD of 0.05 \si{\meter/\px}; 2) our default GSD of 0.01 \si{\meter/\px}; 3) 0.15 \si{\meter/\px}; and 4) 0.20 \si{\meter/\px}.
We create images for the first dataset by cropping image patches of the size equal to the detector's input size of 300~$\times$~300 \si{\px}.
For all other datasets, we cut out the image patches with double/triple/quadruple image size compared to the first dataset, which means that the field-of-view is increasingly larger.
These patches are then down-sampled to 300~$\times$~300 \si{\px}, which increases the GSD.
Since we keep the study site area and the percentage of overlap the same, the number of image patches in the dataset reduces accordingly.
Samples of these four datasets are shown in Figure \ref{fig:art_and_real-gsd} highlighting the loss in detail due to larger GSDs and the increased field-of-view.
For each GSD, we create 1000 artificial images of the same size as the real patches.
The images contain ten vehicles per image, which provides a fixed number of artificial vehicles regardless of GSD.
The test results, achieved with our default feature extraction network ResNet-50, are shown in Table \ref{tab:art_and_real-gsd}.

\begin{table*}
\small
\centering
\caption{Object detection performance of baseline models and models trained with combined images for different ground sampling distances (GSDs) -- A degradation in performance is visible with larger GSDs, but the general pattern is repeated for all GSDs. $N_{\mathrm{r}}$: number of real images,  $N_{\mathrm{a}}$: number of artificial images; $(\cdot)$*: default choice.}
\label{tab:art_and_real-gsd}
\begin{tabular}{ccc|cc|cccc|cccc}
\toprule
     &      & $N_{\mathrm{r}} + N_{\mathrm{a}}$ & \multicolumn{2}{c|}{$N_{\mathrm{r}} + 0$} & \multicolumn{2}{c}{8 + 0} & \multicolumn{2}{c|}{8 + 1000} & \multicolumn{2}{c}{150 + 0} & \multicolumn{2}{c}{150 + 1000} \\
     &      & {} & AP$_{\mathrm{\mu}}\uparrow$ & AP$_{\mathrm{\sigma}}\downarrow$ & AP$_{\mathrm{\mu}}\uparrow$ & AP$_{\mathrm{\sigma}}\downarrow$ & AP$_{\mathrm{\mu}}\uparrow$ & AP$_{\mathrm{\sigma}}\downarrow$ & AP$_{\mathrm{\mu}}\uparrow$ & AP$_{\mathrm{\sigma}}\downarrow$ & AP$_{\mathrm{\mu}}\uparrow$ & AP$_{\mathrm{\sigma}}\downarrow$ \\
GSD [\si{\meter/\px}] & size [\si{\px}] & $N_r$ &                             &                                  &                             &                                  &                             &                                  &                             &                                  &                             &                                  \\
\midrule
0.05 & 300  & 16451 &                       0.956 &                            0.003 &                       0.065 &                            0.109 &                       0.708 &                            0.028 &                       0.869 &                            0.008 &                       0.865 &                            0.009 \\
0.10* & 600  & 2039  &                       0.946 &                            0.008 &                       0.047 &                            0.067 &                       0.740 &                            0.013 &                       0.886 &                            0.008 &                       0.892 &                            0.003 \\
0.15 & 900  & 950   &                       0.911 &                            0.026 &                       0.106 &                            0.040 &                       0.719 &                            0.024 &                       0.852 &                            0.017 &                       0.872 &                            0.004 \\
0.20 & 1200 & 497   &                       0.871 &                            0.004 &                       0.020 &                            0.029 &                       0.581 &                            0.039 &                       0.819 &                            0.012 &                       0.829 &                            0.011 \\
\bottomrule
\end{tabular}

\end{table*}

Overall, the positive effect of the artificial images is observed for all tested GSDs.
However, our experiment shows a trend of reduced overall performance with increasingly larger GSD, which is also reported by \cite{shermeyer2019effects}.
A reason for this is that with larger GSD, details get lost, which are helpful to distinguish vehicles from other structures.
Even though that the smallest GSD of 0.05~\si{\meter/\px} gives the best performance, its margin to 0.1~\si{\meter/\px} GSD is relatively small.
This shows that we do not significantly sacrifice performance by using the latter, which has the benefit of having a smaller dataset and hence allowing faster training.

\paragraph{Comparison with image augmentation}
\label{sec_sub:art_vs_aug}

\begin{figure}[ht]
  \centering

  \includegraphics[width=\linewidth]{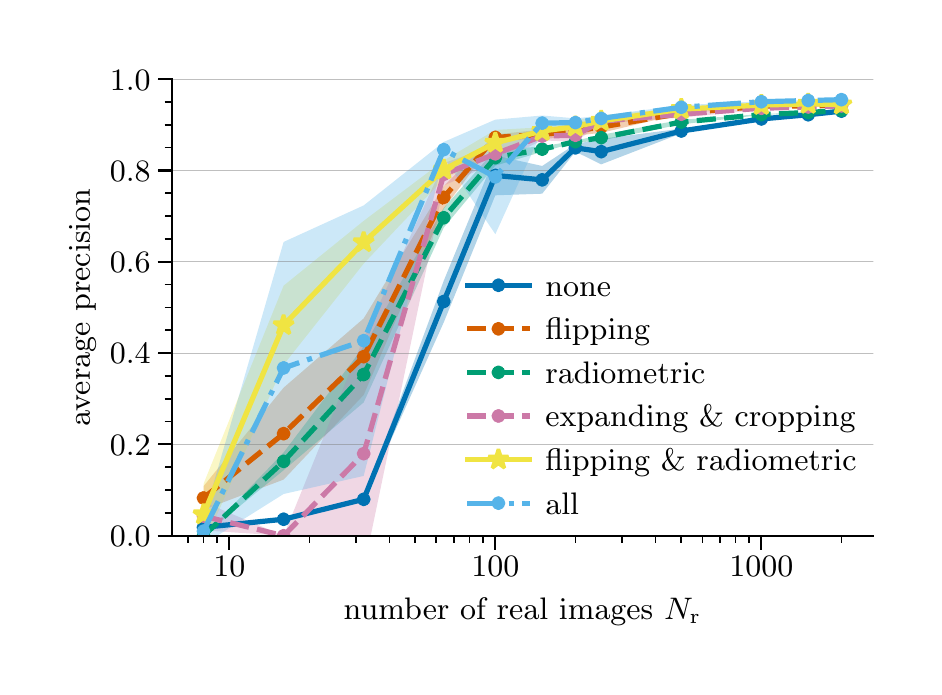}

  \caption{Impact of image augmentation on different sized real datasets -- In the range below 100 images, the different methods' influence is very different and partly negative. Beyond that, the influence between the methods converges, stabilizes, and consistently shows a positive effect.}
  \label{fig:art_vs_aug-augs_no_art}
\end{figure}

\begin{figure}[ht]
  \centering

  \includegraphics[width=\linewidth]{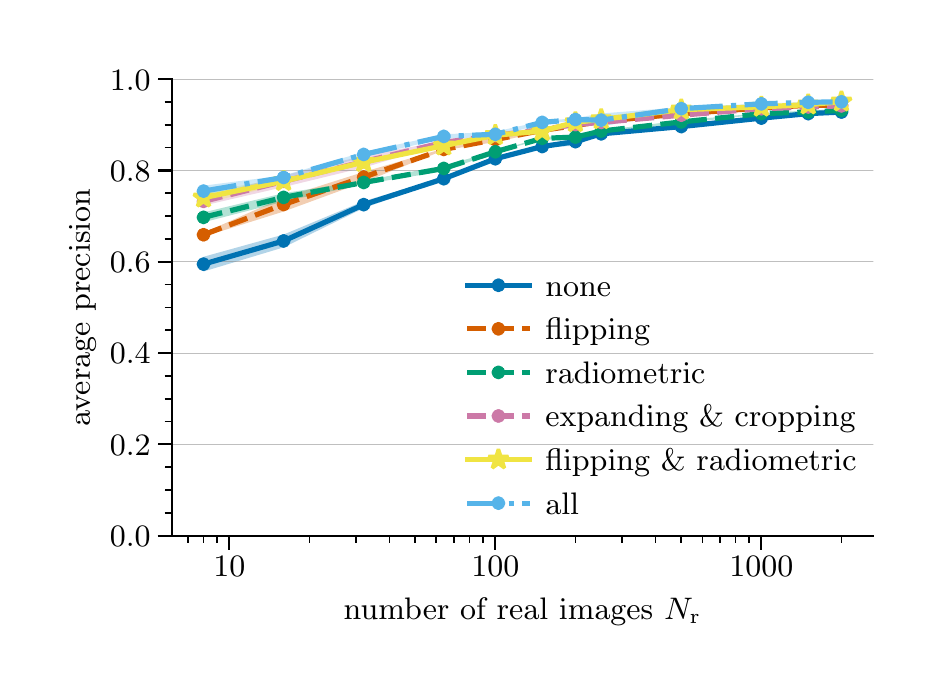}
  
  \caption{Impact of image augmentation on different sized real datasets and additional artificial images -- Adding artificial data significantly boosts and stabilizes the performance regardless of the augmentation method. The differences between individual methods decrease. For the smallest number of real images, there are 0.16 points of AP between the best method combination and the use of no augmentation.}
  \label{fig:art_vs_aug-augs_with_art}
\end{figure}

We stated in Section \ref{sec:related_work} that we regard image augmentation as a complementary approach to increase datasets variability.
To reinforce this, we analyze three augmentation methods, separately and in combination.
These are our two default methods of random flipping of the x and y-axis (flipping) and random alteration of hue, saturation, brightness, and contrast (radiometric), and the more elaborate random expanding and cropping \cite{liu2016ssd} (expanding \& cropping).
We train models for these combinations with the same two groups of datasets from the beginning of this section and show the performance for the ones containing no artificial images in Figure \ref{fig:art_vs_aug-augs_no_art} and the ones with artificial images in Figure \ref{fig:art_vs_aug-augs_with_art}, respectively.

Figure \ref{fig:art_vs_aug-augs_no_art} shows that the results can be divided into two ranges.
The first range spans eight to 100 images, and the second one spans 100 to 2039 images.
The separation is approximately at the dataset size where the artificial images no longer increase performance, underlining our findings from previous experiments (see Section \ref{sec_sub:art_and_real}).  

In the first range, we achieve only negligibly low detection rates without augmentation, and the effect of the augmentations is most significant.
Individually, the simple variants achieve improvements of almost 0.3 points of AP, and in combination more than 0.5 points of AP.
The more complex expanding and cropping augmentation method is unstable alone and destabilizes the performance when used together with the simpler variants.
In total, the differences between all variants are reduced significantly when moving to the second range with 100 and more images.
Here, the performances converge to a range between 0.93 and 0.95~AP.
Simultaneously, the performance variation reduces, which can be seen in the narrower standard deviation bands.
Without any augmentation, however, we reach a marginally smaller accuracy of 0.93~AP.

Figure \ref{fig:art_vs_aug-augs_with_art} shows the results for the combination of artificial and real images using different augmentation methods.
The plot illustrates that the additional artificial images change accuracies significantly.
All performance curves are clearly raised and lie above those of Figure \ref{fig:art_vs_aug-augs_no_art}.
The slope is flatter for all variants, the distances to each other are reduced, and their effect is much more stable.
Besides, the expanding and cropping augmentation no longer results in unstable behavior and shows a lift in performance.
A significant increase in performance is also visible when no augmentation is applied.
However, the difference between this and the best augmentation combination is 0.16 points of AP.

This last observation shows the importance of augmentation and proves our point of seeing it as a useful complementary approach.
The combination of all three methods provides the best performance for large enough datasets, which is in line with the observations in \cite{liu2016ssd}.
We do not use the expanding and cropping augmentation in our training scheme for all other experiments due to the instability with small datasets.

\subsection{Details of the artificial image generator}
\label{sec_sub:art-ablation}

To better understand the observed performance gap, we look at different components of the generator to analyze which aspects of the images are essential.
We incrementally add the components described in Section \ref{sec_sub:framework-art} and visualize them in Figure \ref{fig:art-gen-pipeline}.
We further vary the deformation and background noise hyper-parameters and create a separate dataset for each of these variations.
In this scenario, we do not use real images for training in contrast to the previous experiment.
This extreme case shows a clearer picture of the effects, which additional real images would otherwise hide.
We train models for each of these datasets and present the resulting average precision in Table \ref{tab:artificial_variants}.

\begin{table*}
    \centering
    
    \caption{Impact of the image generator components on the detection performance -- It is apparent that structure and variation in both vehicles and background help to improve the effectiveness of the images. The deformations should only be slight. The same applies to background noise.}
    \label{tab:artificial_variants}
    \begin{tabular}{ccccccc}
    \toprule
    border &      partially depicted & deform & fine noise & rough noise &   AP$_{\mathrm{\mu}}\uparrow$ &    AP$_{\mathrm{\sigma}}\downarrow$ \\
    \midrule
      body &             &        &             &             &  0.521 &  0.021 \\
     black &             &        &             &             &  0.553 &  0.014 \\
     black &  \checkmark &        &             &             &  \textbf{0.581} &  0.009 \\
     \midrule
     black &  \checkmark &      5 &             &             &  \textbf{0.627} &  0.009 \\
     black &  \checkmark &     10 &             &             &  0.610 &  0.005 \\
     black &  \checkmark &     20 &             &             &  0.561 &  0.006 \\
     \midrule
     black &  \checkmark &      5 &  \checkmark &             &  0.660 &  0.008 \\
     black &  \checkmark &      5 &  \checkmark &           5 &  \textbf{0.683} &  0.004 \\
     black &  \checkmark &      5 &  \checkmark &          10 &  \textbf{0.686} &  0.004 \\
     black &  \checkmark &      5 &  \checkmark &          15 &  0.674 &  0.010 \\
     black &  \checkmark &      5 &  \checkmark &          20 &  0.662 &  0.014 \\
    \bottomrule
    \end{tabular}
\end{table*}

The best configuration reaches about 0.68~AP, whereas, in comparison, baseline models reach 0.64 and 0.80~AP with 32 and 64 real images, respectively.
This shows that, although these models are only trained with artificial images, they can successfully detect objects in real images.
The table shows that each component improves the data quality and improves the average precision.
However, a few aspects are noteworthy:
If we are not using black vehicle outlines and draw them in body color, the performance decreases by 0.032 points of AP, which shows that the details of the objects matter and that there is a lower limit for simplicity.
The addition of partially depicted vehicle adds 0.028 points of AP, which helps the fact that the dataset contains multiple vehicles partially covered by trees or buildings.
Including a slight deformation helps increase the variety of size and shape of the vehicles and gives a rise of 0.046 points of AP.
On the other hand, the deformation should not be too large as this results in unrealistic shapes and aspect ratios that harm the model.
The most interesting aspect will be the improvement when noise is added to the background, which gives almost 0.06 points of AP for both variants of noise combined.
The background differentiates the artificial and real images the most, and real background typically can contain lots of structure.
Our notion is that models, which have not been trained with a background signal, can confuse these background structures with relevant objects.
This is why adding a bit of structure to the hitherto blank background results in a small step towards more realistic backgrounds and reduces confusion.

\subsection{The importance of image composition}
\label{sec_sub:art-composition}

To substantiate the last section's observation, we investigate more in-depth into the effect the background and the image composition has on the detection performance.
We will show the qualitative results of models trained with real and artificial data, which makes it evident that the background structure is important. 
In detail, we compose various image patches by combining artificial and real vehicles and background, respectively, to analyze the influence of artificial and real image components on the performance.

\begin{figure}
    \centering
    
    \begin{subfigure}[t]{0.24\linewidth}
    \includegraphics[width=\linewidth]{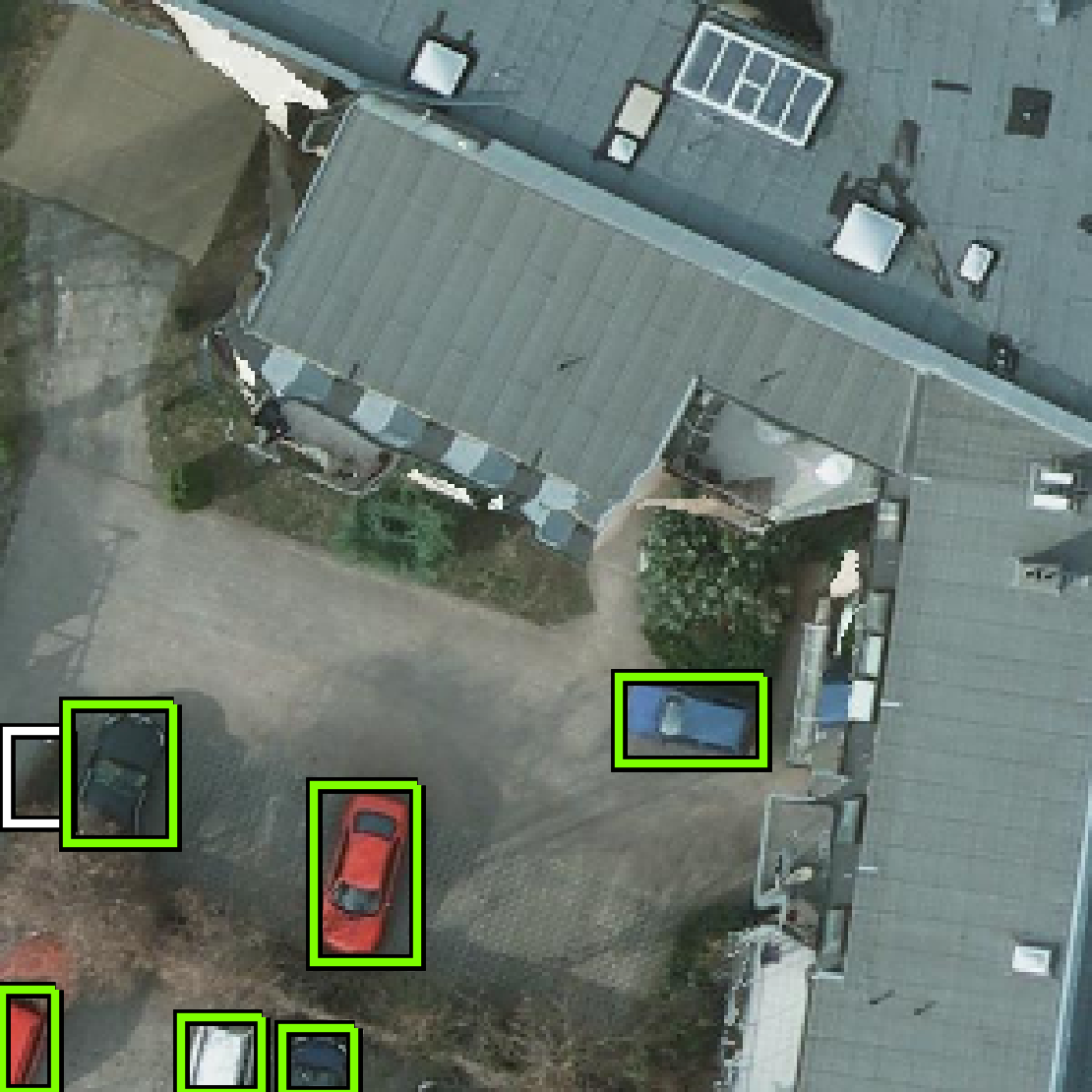}
    \end{subfigure}
    \hfill
    \begin{subfigure}[t]{0.24\linewidth}
    \includegraphics[width=\linewidth]{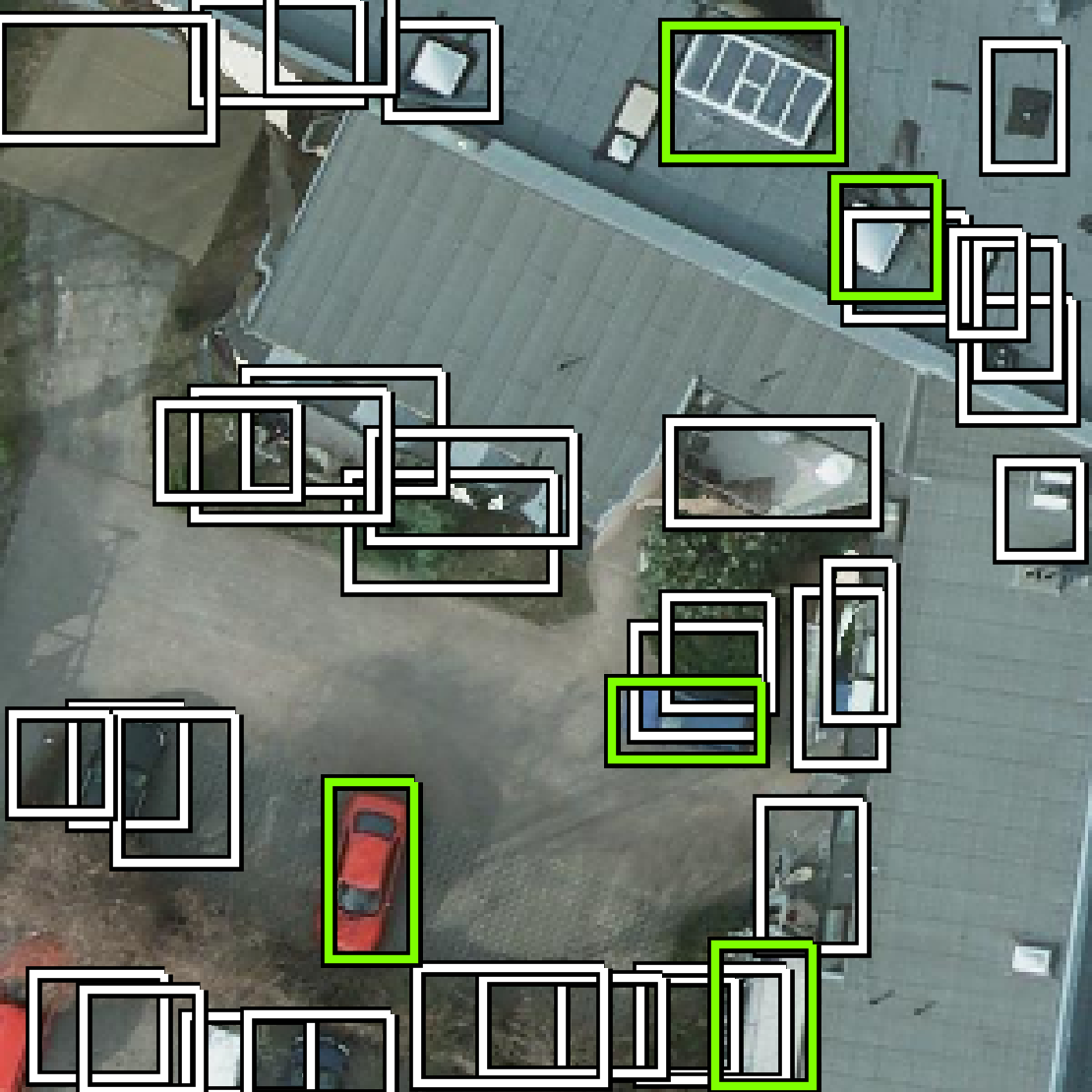}
    \end{subfigure}
    \hfill
    \begin{subfigure}[t]{0.24\linewidth}
    \includegraphics[width=\linewidth]{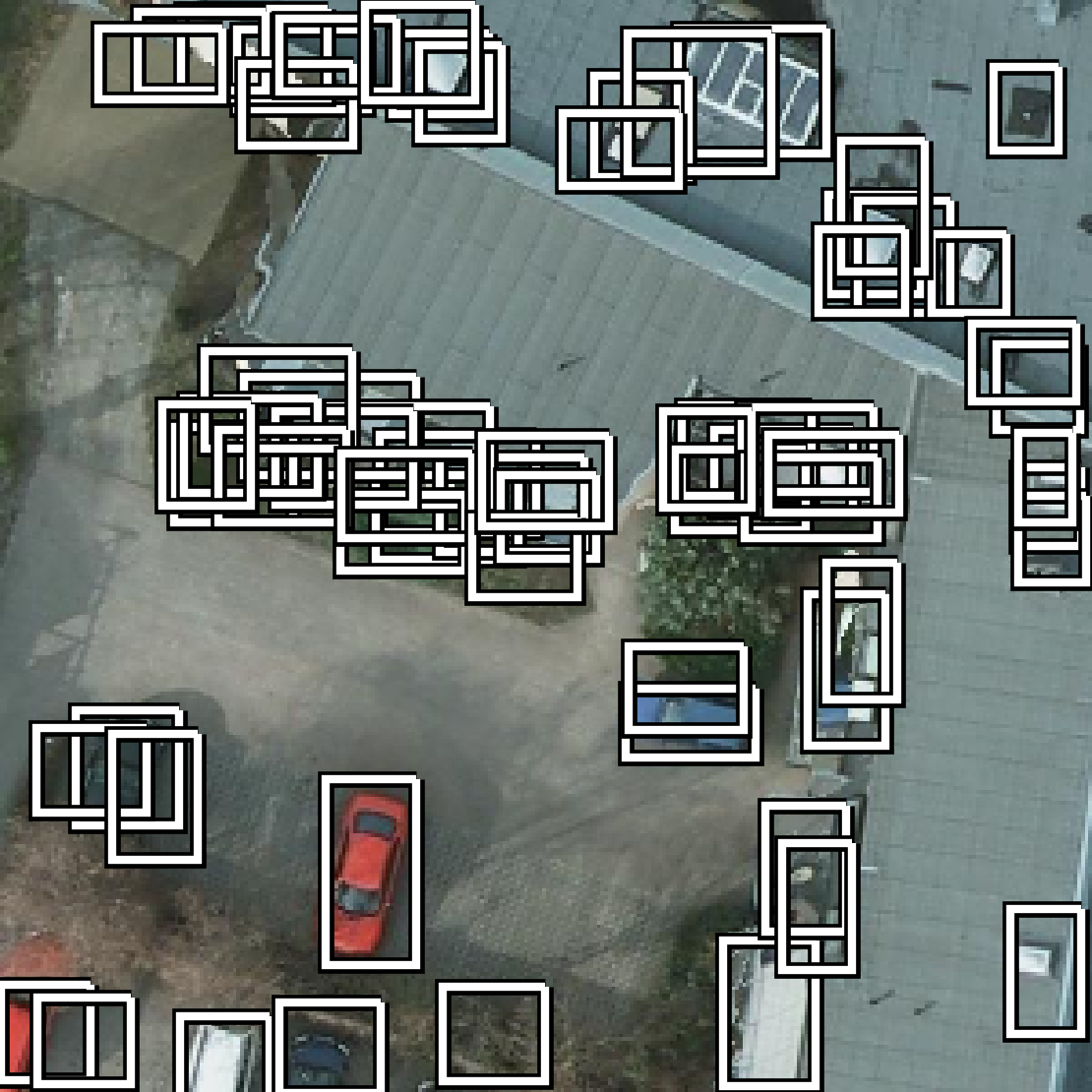}
    \end{subfigure}
    \hfill
    \begin{subfigure}[t]{0.24\linewidth}
    \includegraphics[width=\linewidth]{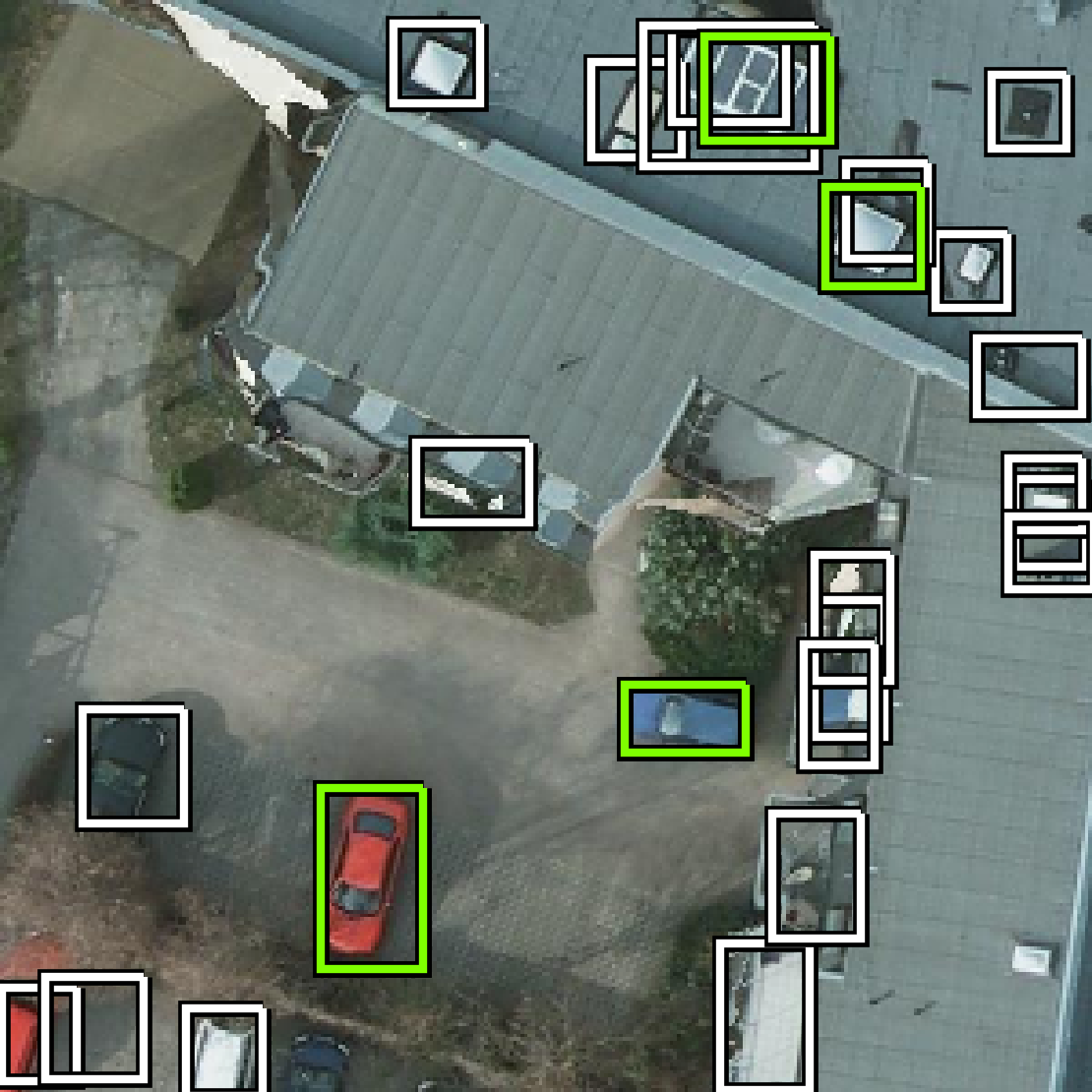}
    \end{subfigure}
    
    \medskip
    
    \begin{subfigure}[t]{0.24\linewidth}
    \includegraphics[width=\linewidth]{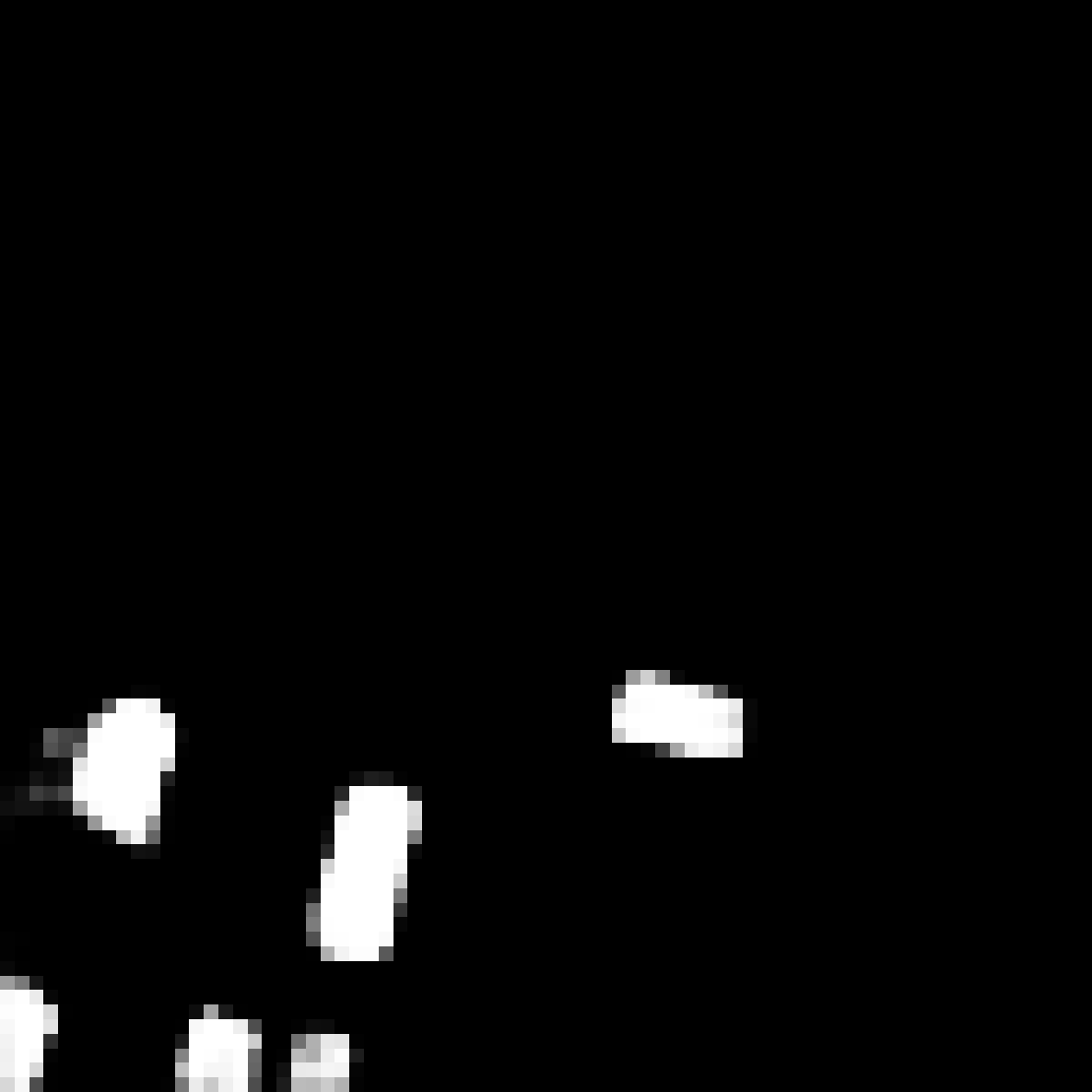}
    \caption{\raggedright real images (0.952~AP)} 
    \label{fig:art-composition-predictions-real}
    \end{subfigure}
    \hfill
    \begin{subfigure}[t]{0.24\linewidth}
    \includegraphics[width=\linewidth]{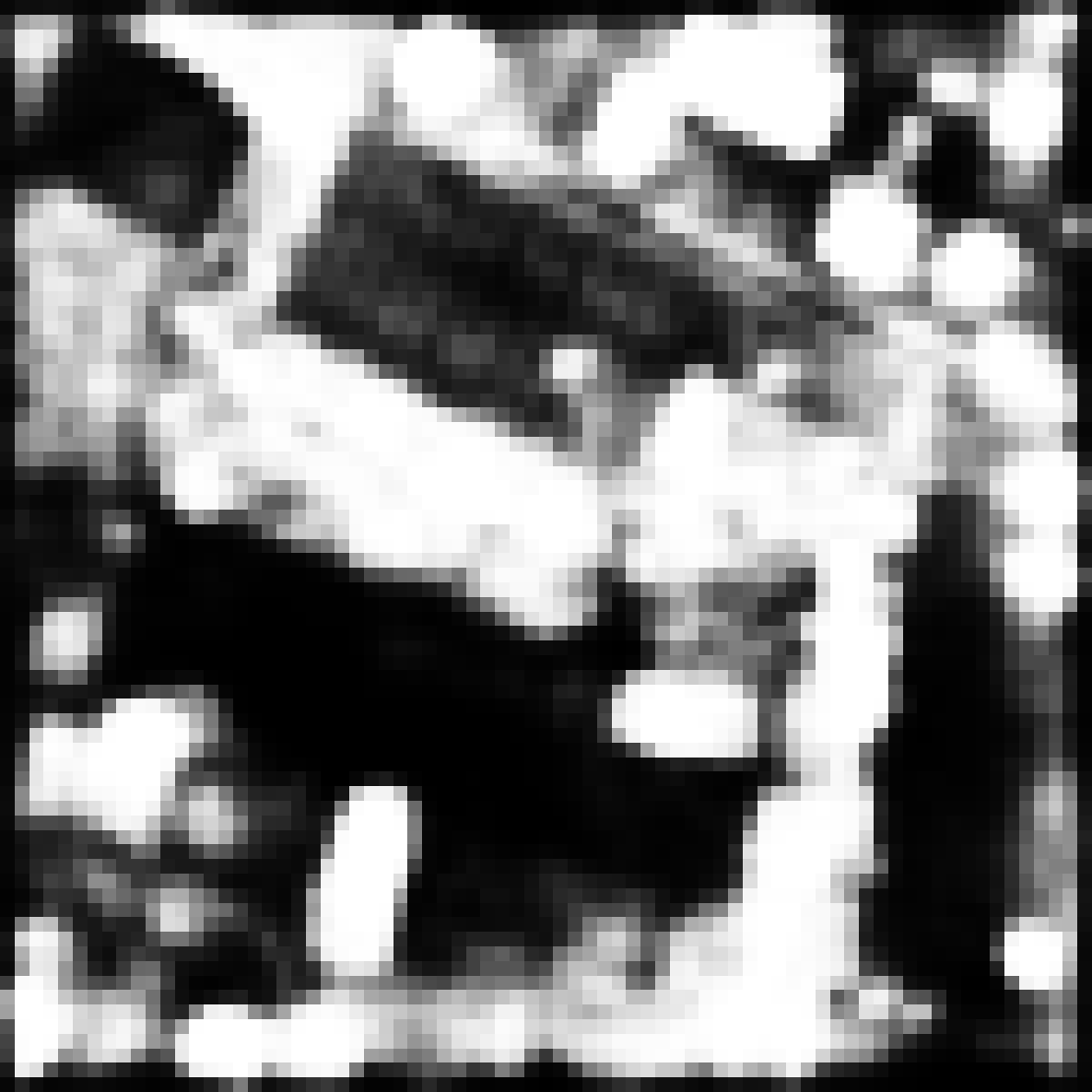}
    \caption{\raggedright artificial images with a plain background (0.627~AP)} 
    \label{fig:art-composition-predictions-art_nobg}
    \end{subfigure}
    \hfill
    \begin{subfigure}[t]{0.24\linewidth}
    \includegraphics[width=\linewidth]{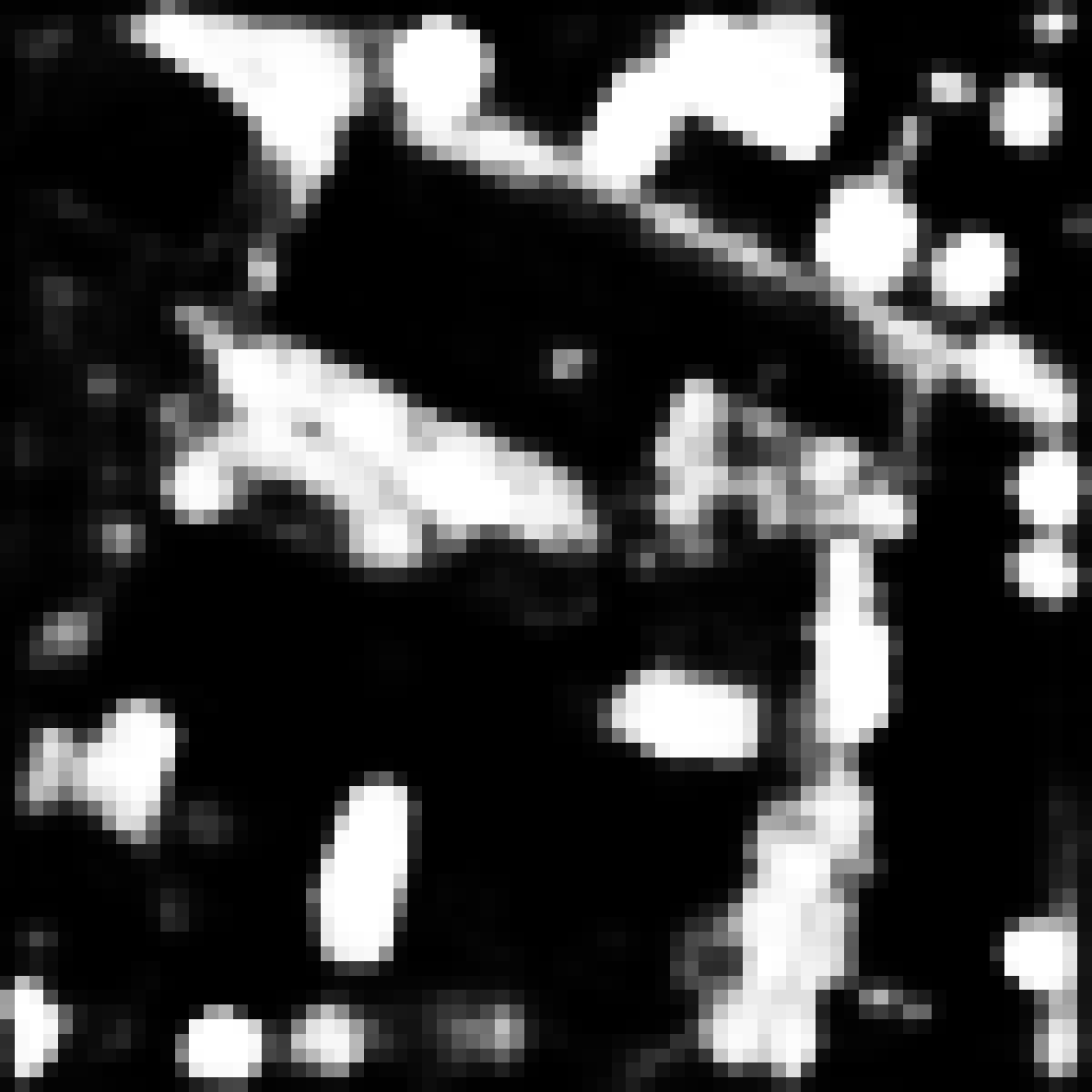}
    \caption{\raggedright artificial images with background noise (0.686~AP)} 
    \label{fig:art-composition-predictions-art_noise}
    \end{subfigure}
    \hfill
    \begin{subfigure}[t]{0.24\linewidth}
    \includegraphics[width=\linewidth]{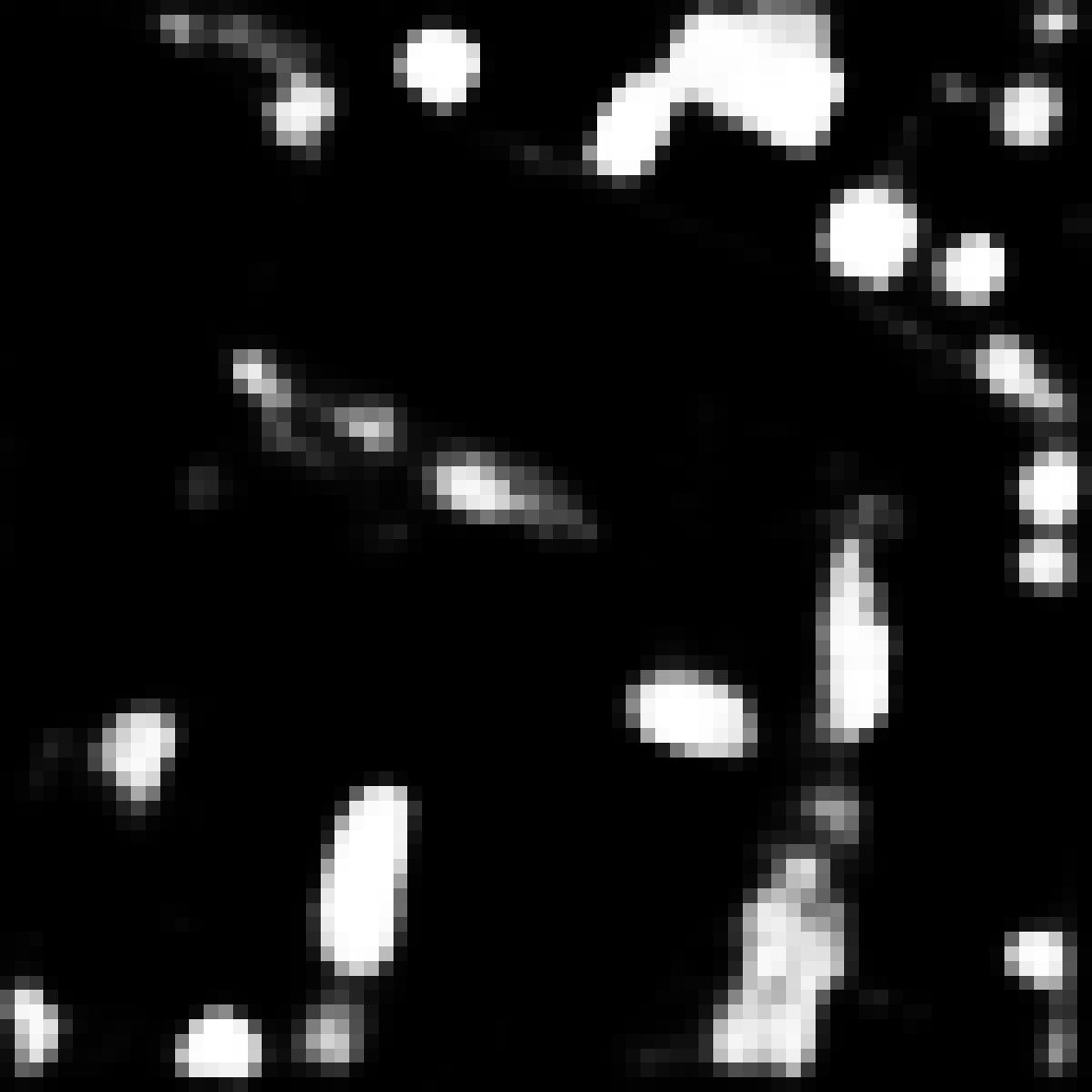}
    \caption{\raggedright artificial images with background noise and eight real images (0.736~AP)} \label{fig:art-composition-predictions-art_noise_8real}
    \end{subfigure}
    
    \caption{A model trained with real images (left) in comparison with models trained with different artificial images (top: predicted bounding boxes, green: confidence $>$ 0.5, white: confidence $>$ 0.1; bottom: semantic activation map) -- In contrast to the left baseline model, the other models show significant false-positives for both types of predictions, but, a clear trend in improvement is visible in the semantic activation with more elaborate background and addition of some real images.}
    \label{fig:art-composition-predictions}
\end{figure}

\paragraph{The effect of background}
Since our results indicate that adding noise to artificial background gives performance improvements, we provide a qualitative accuracy assessment of the predictions for a more detailed understanding.
We train and compare models with four of the previous datasets: 1) containing 2039 real images; 2) 1000 artificial images with a plain background; 3) 1000 artificial images with a noisy background, and 4) the same images as before extended by eight real images.
Figure \ref{fig:art-composition-predictions} shows the predictions these models make on a single image of the test dataset.
We show the bounding boxes in the top row and the semantic segmentation branch's activation maps in the bottom row.
We use the same confidence thresholds and coloring as in Figure \ref{fig:art_and_real-samples}.

The baseline model on the left (Fig. \ref{fig:art-composition-predictions-real}) predicts bounding boxes with high accuracies and produces a high-quality semantic activation map.
The single white box is a false-positive as it passes the evaluation confidence threshold but stays below 0.5.
The second model (Fig. \ref{fig:art-composition-predictions-art_nobg}) produces many bounding boxes that pass the lower threshold and result in a larger false-positive rate.
However, only for two vehicles, an actual high confidence prediction exists.
A significant number of false-positive pixels are contained in the activation map, mostly located in structured image regions but comparatively lower in more homogeneous areas of the image.
We assume this is because the artificial images' homogeneous blank backgrounds only let the model learn to differentiate structure from the structure-less image content.
Nevertheless, a strong signal for the image areas with actual vehicles is also present.
For the third model (Fig. \ref{fig:art-composition-predictions-art_noise}), a small improvement for bounding boxes in the vicinity of vehicles can be seen, but the model still confuses a lot of background structure with vehicles.
However, a reduction of false-positive pixels in the activation map is visible. 
The added noise on the image's background is responsible for this, as it is the only difference between the two datasets.
The last model (Fig. \ref{fig:art-composition-predictions-art_noise_8real}) has been trained with additional eight real images, which have a notable impact on the model's predictions.
Even though not reflected in the higher confidence boxes, the false-positive rate for both bounding boxes and semantic segmentation decreases largely.
This is a clear hint that structure from the additional real images, of which most is background rather than vehicles, is crucial for the model.
Comparing the semantic activation maps of all three models trained with artificial images, it is apparent that the pixel activation for actual vehicles is solid and stable.
This can also be observed for most of the remaining false-positive image areas, mostly rectangular features, that resemble artificial vehicles' features.
Features like these are commonly called hard negatives due to the described issues models have.
The added background information helps eradicate the majority of false-positives and sharpens the vehicle areas but does not significantly raise more true positives.

\begin{figure}
    \centering
    
    \begin{subfigure}[t]{0.24\linewidth}
    \includegraphics[width=\linewidth]{scene_roughnoise.png}
    \caption{\raggedright artificial vehicles and artificial background} 
    \label{fig:art_to_real-variants-art_best}
    \end{subfigure}
    \hfill
    \begin{subfigure}[t]{0.24\linewidth}
    \includegraphics[width=\linewidth]{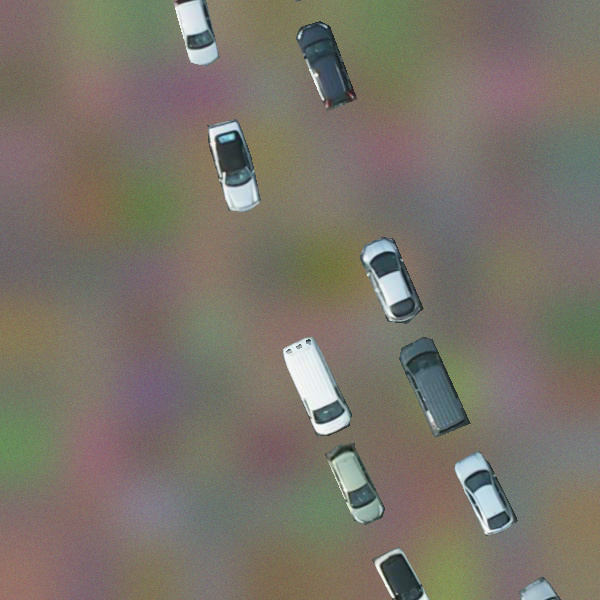}
    \caption{\raggedright real vehicles and artificial background} 
    \label{fig:art_to_real-variants-real_fakeback}
    \end{subfigure}
    \hfill
    \begin{subfigure}[t]{0.24\linewidth}
    \includegraphics[width=\linewidth]{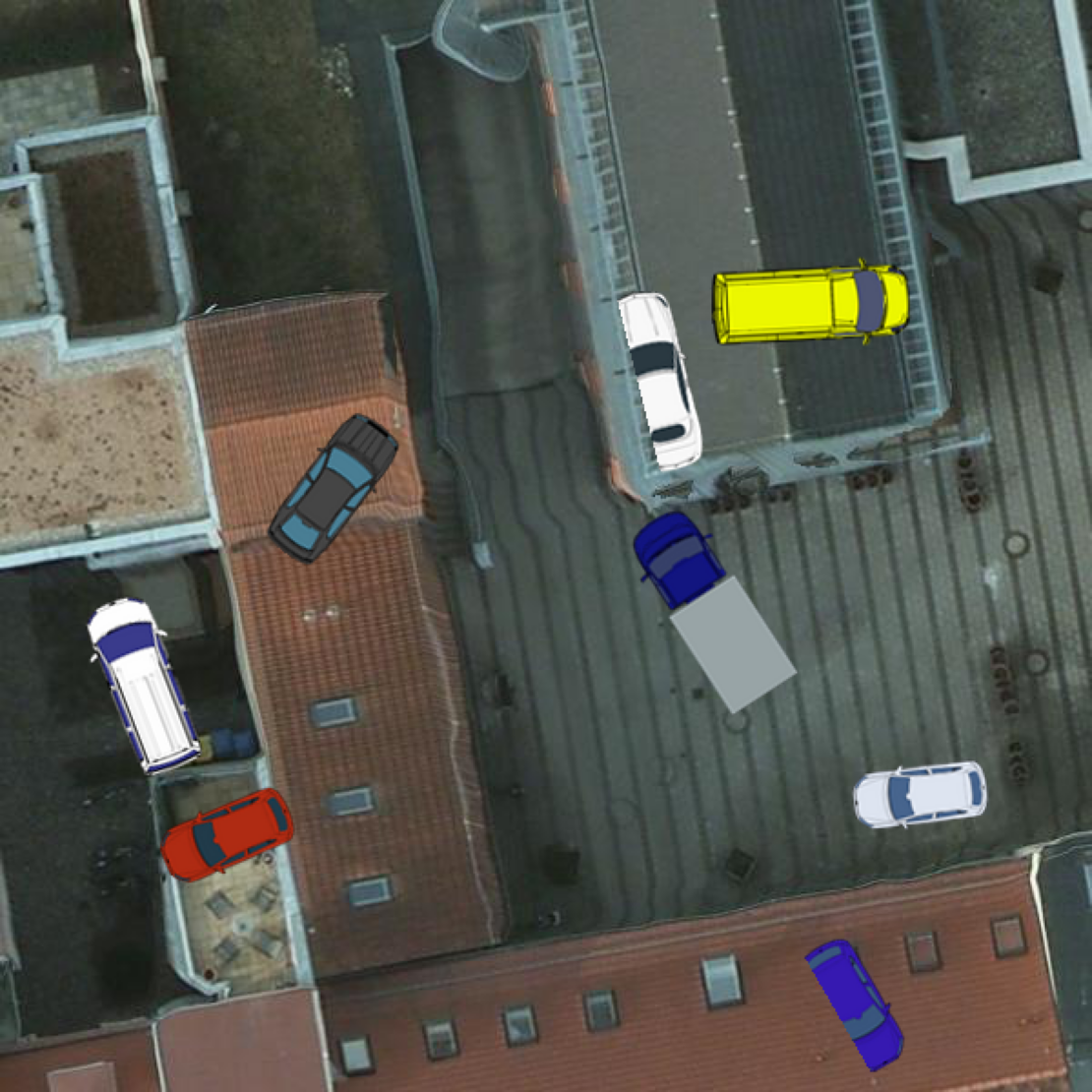}
    \caption{\raggedright artificial vehicles and real random background} \label{fig:art_to_real-variants-art_randback}
    \end{subfigure}
    \hfill
    \begin{subfigure}[t]{0.24\linewidth}
    \includegraphics[width=\linewidth]{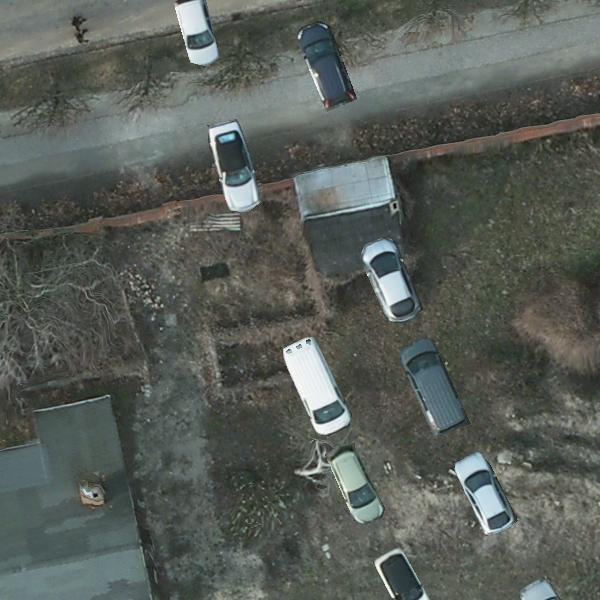}
    \caption{\raggedright real vehicles and real random background} 
    \label{fig:art_to_real-variants-real_randback}
    \end{subfigure}
    
    \caption{Samples of mutually combined real and artificial vehicles and background -- The difference in degree of detail is apparent between \subref{fig:art_to_real-variants-art_best}) and \subref{fig:art_to_real-variants-real_fakeback}), the vehicles in \subref{fig:art_to_real-variants-art_randback}) look foreign to the background and the ignored semantics stand out, which is also the case for \subref{fig:art_to_real-variants-real_randback}).}
    \label{fig:art_to_real-variants}
\end{figure}

\paragraph{Image composition}
In order to analyze our previous findings in more detail, we assess the performance of various image compositions.
For this, we create images by combining artificial vehicles and backgrounds with their real counterparts in all four combinations.
We use image patches from the Potsdam dataset for real backgrounds, which do not contain any vehicles.
The distribution of scenery is similar between images without vehicles and images with vehicles so that the resulting domain shift is minimal.
We sample from the set of images that only show background and add either real or artificial vehicles while neglecting the semantics between foreground and background. 
Samples of these four datasets are shown in Figure \ref{fig:art_to_real-variants} and the resulting average precision of models trained with these datasets are shown in Table \ref{tab:art_to_real-variants}.

\begin{table}
    \centering
    
    \caption{Average precision (AP) of models trained with the datasets containing mutually combined real and artificial vehicles and background.}
    \label{tab:art_to_real-variants}
    \begin{tabular}{cccc}
    \toprule
     vehicles &   background &  AP$_{\mathrm{\mu}}\uparrow$ &  AP$_\mathrm{\sigma}\downarrow$ \\
    \midrule
     artificial &         artificial &                0.686 &                 0.004 \\
     real &         artificial &                0.733 &                 0.008 \\
     artificial &  real random &                0.726 &                 0.009 \\
     real &  real random &                0.890 &                 0.006 \\
     real &         real &                0.932 &                 0.009 \\
    \bottomrule
    \end{tabular}
\end{table}

First, we compare the results for the dataset with only artificial images (see Fig. \ref{fig:art_to_real-variants-art_best} for an example) and the dataset with real vehicles on an artificial noisy background (see Fig. \ref{fig:art_to_real-variants-real_fakeback} for an example). Since both datasets do not contain detailed background information, the average precision gap can be attributed to the differences between the vehicles.
As the difference in accuracy is less than 0.05 points of AP, we suppose that our relatively simple artificial vehicles already contain useful, relevant features for training a model for vehicle detection.
More photo-realistic vehicles and a more fine-tuned selection of blueprints can potentially close the small accuracy gap.

Comparing the results obtained from the dataset with standard artificial images (Fig. \ref{fig:art_to_real-variants-art_best}) and the results obtained with the dataset containing artificial vehicles on real background (Fig. \ref{fig:art_to_real-variants-art_randback}), we analyze whether real background can significantly improve performance.
Since the gain in accuracy is 0.04, which is significant but relatively low, our experiments indicate that using real background information alone will not significantly improve the artificial images' usefulness.  
For a human observer, it is obvious that the artificial vehicles look like foreign objects. 
Therefore we suspect that a more harmonized combination \cite{bhattad2020cut} and using domain adaption \cite{sun2016deep,koga2020method} could help to improve performance.

Neglecting the semantics between foreground and background can result in incorrect placement of vehicles, as they can randomly be placed on houses or other structures, where they, in reality, typically would not appear.
Therefore, in our last sub-experiment, we analyze the importance of these semantics and compare the datasets' results with real vehicles on either matching or randomized real backgrounds (see Fig. \ref{fig:art_to_real-variants-real_randback}).
The average precision of 0.89 obtained from the latter dataset is marginally smaller than the average precision of 0.93 obtained from the baseline dataset. 
The difference between these values can be accounted to the loss of semantics. 

To sum up, the experiments show that our artificially designed vehicles already contain important features to improve the performance of small real datasets notably and that semantics do not play a significant role.
The latter finding is promising since otherwise, semantically correct images would need to be created, which is challenging due to arbitrarily complex geometric conditions.
Finally, we found that creating images with artificial vehicles and real backgrounds instead of a simple noisy background does not largely improve the performance.
We expect that our observations and findings can point towards important aspects of the image composition and lead to improvements of artificial data that can further reduce the gap towards real data and boost their positive effect on small real datasets.

\section{Conclusion and future work}
\label{sec:conclusion_future}
In this work, we addressed the limited data availability for vehicle detection in aerial imagery and provided insights into the gain from adding artificial imagery to the training dataset.
The core of our framework consists of a RetinaNet detector, which is modified to be particularly suitable for our task.
We utilized a generative approach to create artificial images from simplified 2D CAD drawings and analyzed their impact on the performance. 
In order to follow our objective of providing a better understanding of the effect of a lack of data and to what extent artificial data can compensate this effect, for our experiments, we simulated a small dataset based on the ISPRS Potsdam dataset.
Experiments with additional artificial images, based on eight 2D CAD blueprints, reveal that more stable results and a significantly higher performance rate can be achieved.
For example, we have seen a gain of about 0.7 points of average precision when using 8 real images with additional 1000 artificial images.
Moreover, we observed that models trained with only artificial images are able to detect vehicles in real images, where we reached similar gains with 0.68 average precision points when using 1000 artificial images. 

Our key findings of our work are:
\begin{itemize}
    \item datasets containing only a very few samples have a severe impact on the achievable detection performance;
    \item for small datasets, adding new samples has a large impact that decreases as the dataset gets larger;
    \item adding simple artificial images can alleviate a lack of real data and significantly improve object detection performance;
    \item the use of artificial images showed to be beneficial regardless of network capacity and ground sampling distance;
    \item for small datasets enhanced with images containing artificial objects, the remaining performance gap to large real datasets cannot be closed by using real backgrounds;
    \item semantics plays a subordinate role when real or artificial objects are combined with different backgrounds; 
    \item adding artificial images is a complementary approach to image augmentation.
\end{itemize}

Our work builds the basis for promising future directions:
Without the need to collect and annotate real data, this can be a great advantage for object detection, especially for classes that do not frequently occur in reality.
If the combination of real background and artificial vehicles can be successfully implemented, an artificial generator can provide a high degree of freedom in image creation.
A promising direction is to disentangle the underlying factors of variation of foreground and background at the image level and use them for an improved generation process. 
Instead of using all combinations of background and foreground, relevant factors could be simulated and their combination modeled in the network. Especially for remote sensing, where various sceneries may occur, this would be a great advantage.
We also see a promising direction in finding out more about which elements in real-world objects lead to a significant performance increase. Knowing which elements are crucial, the effort of artificial data generation can be kept low with a maximum possible performance increase.

\section*{Acknowledgement}
The authors would like to thank the DVGW -- German Technical and Scientific Association for Gas and Water -- for funding this work as part of the research project G201819 - Antonia.

\bibliography{references}

\end{document}